\documentclass{article} 
\usepackage{iclr2027_conference,times}

\usepackage{amsmath,amsfonts,bm}

\def\eqref#1{equation~\ref{#1}}

\def\1{\bm{1}}

\DeclareMathAlphabet{\mathsfit}{\encodingdefault}{\sfdefault}{m}{sl}
\SetMathAlphabet{\mathsfit}{bold}{\encodingdefault}{\sfdefault}{bx}{n}

\usepackage{hyperref}
\usepackage{url}
\usepackage{graphicx}
\usepackage{float}
\usepackage{booktabs}
\usepackage{placeins}
\usepackage{subcaption}
\usepackage{titlesec}

\titlespacing{\section}
{0pt}{8pt}{4pt}

\titlespacing{\subsection}
{0pt}{6pt}{3pt}
\usepackage{titlesec}

\title{What Limits Recursive Reasoning Models: Optimization, Architecture and Test-Time Scaling}

\author{
\makebox[\textwidth][c]{%
\bfseries
\begin{tabular}{@{}ccc@{}}
Yuliana Shakhvalieva\thanks{Corresponding author.} &
Dmitrii Kharchev &
Viacheslav Bezrukov
\end{tabular}
}
\\[-3pt]
\makebox[\textwidth][c]{%
\bfseries
\begin{tabular}{@{}cccc@{}}
Inessa Fedorova &
Dmitry Bocharov &
Ivan Oseledets &
Valerii Ternovskii
\end{tabular}
}
\\[4pt]
\makebox[\textwidth][c]{RND NLP, DAIMLD, Russian Federation}
\\[-1pt]
\makebox[\textwidth][c]{\texttt{ysshakhvalieva@daimld.tech}}
}

\iclrfinalcopy 

\begin{document}

\maketitle

\begin{abstract}
Recursive reasoning models apply a small shared Transformer block many times to refine a latent state. This gives them large effective depth with few parameters and makes them strong on algorithmic tasks. Such compact solvers are natural candidates for tools that an LLM can call on narrow algorithmic subproblems. However, existing models such as HRM, TRM and URM differ in architecture, gradient propagation and training procedure simultaneously. This makes it hard to tell what drives their performance, and their optimization is still poorly understood and often unstable. In this work we address both of these gaps. First, we study these questions under a unified experimental pipeline spanning six algorithmic domains. Individual controlled ablations are performed on representative domains, while the resulting recipe is evaluated across the full suite. The study reveals a surprisingly simple recipe for stable and generalizable recursive reasoning: an intermediate gradient horizon, large physical batches and controlled updates of the recurrent state. An explicit hierarchical architecture is not needed. Second, we combine these findings into a stable 13.6M-parameter model that achieves the strongest overall performance among the evaluated recursive baselines, with particularly large gains on out-of-distribution generalization. It raises Arithmetic OOD accuracy to 71.2\%, from 36.2\% for the strongest baseline, while reaching 98.41\% on Sudoku and 59.5\% pass@2 on ARC-AGI-1. Our results show that, within the recursive architectures studied here, performance depends strongly on how recurrence is optimized and stabilized. More broadly, it shows how AI systems can be improved by optimizing their components one at a time.
\end{abstract}
\begin{figure}[H]
    \centering
    \makebox[\textwidth][c]{%
        \includegraphics[
            width=\textwidth,
            trim=8 5 10 5,
            clip
        ]{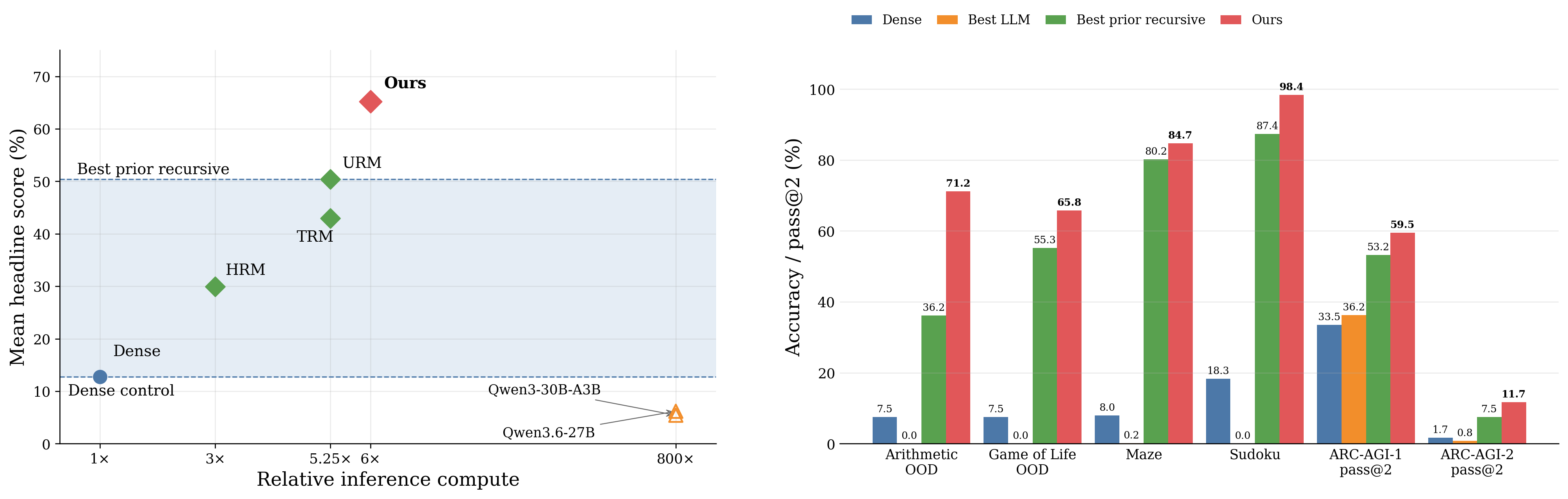}
    }
    \label{fig:teaser}
    \caption{
Recursive reasoning performance is determined by both recurrent computation and its optimization. 
We systematically study the factors that govern recursive models: gradient propagation through the recurrent trajectory, recurrent-state stabilization, architectural choices, and test-time computation. 
Our final 13.6M-parameter model combines the most effective components and achieves strong performance across algorithmic domains while improving out-of-distribution generalization.
}
\end{figure}

\section{Introduction}
\label{sec:introduction}

Recursive reasoning models repeatedly apply a small shared network to a latent
state, trading parameter count for computational depth. Recent models such as
the Hierarchical Reasoning Model (HRM), Tiny Recursive Model (TRM), and
Universal Reasoning Model (URM) show that this approach can solve symbolic
algorithmic tasks with models of only a few million parameters
\citep{wang2025hrm,jolicoeurmartineau2025trm,gao2025urm}.

It is much less clear \emph{why} these models work. HRM, TRM, and URM change
recurrent architecture, gradient propagation, optimization, and regularization
simultaneously, making it difficult to attribute improvements to individual
design choices. Moreover, repeated application of shared parameters creates
optimization challenges that differ from standard feed-forward Transformers.

We study these questions under a unified experimental pipeline. We retrain
recursive baselines with matched data and evaluation, varying one architectural
or optimization factor at a time. Beyond standard benchmarks (Sudoku, Maze,
and ARC-AGI), we introduce Game of Life and Arithmetic with controlled
out-of-distribution splits to test whether learned computation generalizes
beyond the training distribution.

Our study yields three findings. First, optimization choices account for a
large fraction of the performance variation observed in our study: gradient
propagation has an interior optimum, and large physical batches outperform
gradient accumulation at the same effective size. Second, stable recurrent
computation requires controlling latent-state updates; recurrent-state
stabilization is critical for reliable training. Third, additional capacity
does not translate uniformly into better reasoning: explicit hierarchy,
larger recurrent blocks, and additional test-time
computation provide domain-dependent benefits.

Combining these findings gives a stable 13.6M-parameter recursive reasoner that
achieves the strongest results among the evaluated recursive baselines on
nearly all metrics, including 71.16\% on Arithmetic OOD, 98.41\% on Sudoku,
and 59.50\% pass@2 on ARC-AGI-1.

Our contributions are:

\begin{itemize}
    \item \textbf{A controlled ablation study of recursive reasoning.}
    We isolate architectural and optimization choices of HRM/TRM/URM-style
    models under a common training and evaluation pipeline across six domains.

    \item \textbf{Explicit tests of algorithmic generalization.}
    We introduce Game of Life and Arithmetic datasets with controlled
    out-of-distribution splits over computational horizon, input structure, and
    target range.

    \item \textbf{A practical recipe for stable recursive optimization.}
    We identify an intermediate gradient horizon, large physical batches, and
    controlled recurrent-state updates as the key ingredients in our setting.

    \item \textbf{Scaling limits of the recipe.}
    We show that explicit hierarchy, larger recurrent blocks, and additional test-time computation provide domain-dependent benefits.
\end{itemize}

\section{Related Work}
\label{sec:related-work}

\paragraph{Latent recurrent computation.}
Reusing a shared transformation across multiple iterations provides additional
computational depth without increasing parameter count. Looped Transformers
demonstrate this principle on algorithmic reasoning and length generalization
\citep{saunshi2025latent,fan2025looped}, while recurrent-depth language models
extend it to test-time computation in larger language models
\citep{geiping2025scaling}. Related approaches make recurrent depth adaptive
across tokens \citep{bae2025mixture} or perform search directly in a continuous
latent program space \citep{macfarlane2025searching}. These results establish
latent iteration as a useful computational primitive, but leave open how such
recurrence should be optimized and stabilized.

\paragraph{Small recursive reasoners.}
HRM introduced a two-timescale architecture with separate fast and slow
recurrent modules, truncated credit assignment, and adaptive computation
\citep{wang2025hrm}. TRM subsequently showed that much of this architectural
structure can be removed: it shares parameters across recurrent levels,
extends the differentiated portion of the trajectory, and uses weight averaging
\citep{jolicoeurmartineau2025trm}. URM retains the shared-module design while
adding local convolutional mixing and a different truncated-backpropagation
scheme \citep{gao2025urm}. Subsequent analyses further question whether
hierarchical structure or learned halting is essential
\citep{ge2025perspectives,movahedi2026fixedpoint}, and alternative
interpretations view recurrent updates as policy-improvement operations
\citep{asadulaev2026latent}.

These models improve along several axes simultaneously. In particular, HRM,
TRM, and URM change architecture, gradient horizon, optimizer, regularization,
and evaluation setup together, making it difficult to identify which factor
drives their gains. Our work is complementary: rather than proposing another
independent architecture, we place these choices in a shared experimental
framework and vary them one at a time.


\section{Data}
\label{sec:data}

\begin{figure}[!htbp]
    \centering
    \includegraphics[width=0.98\linewidth]{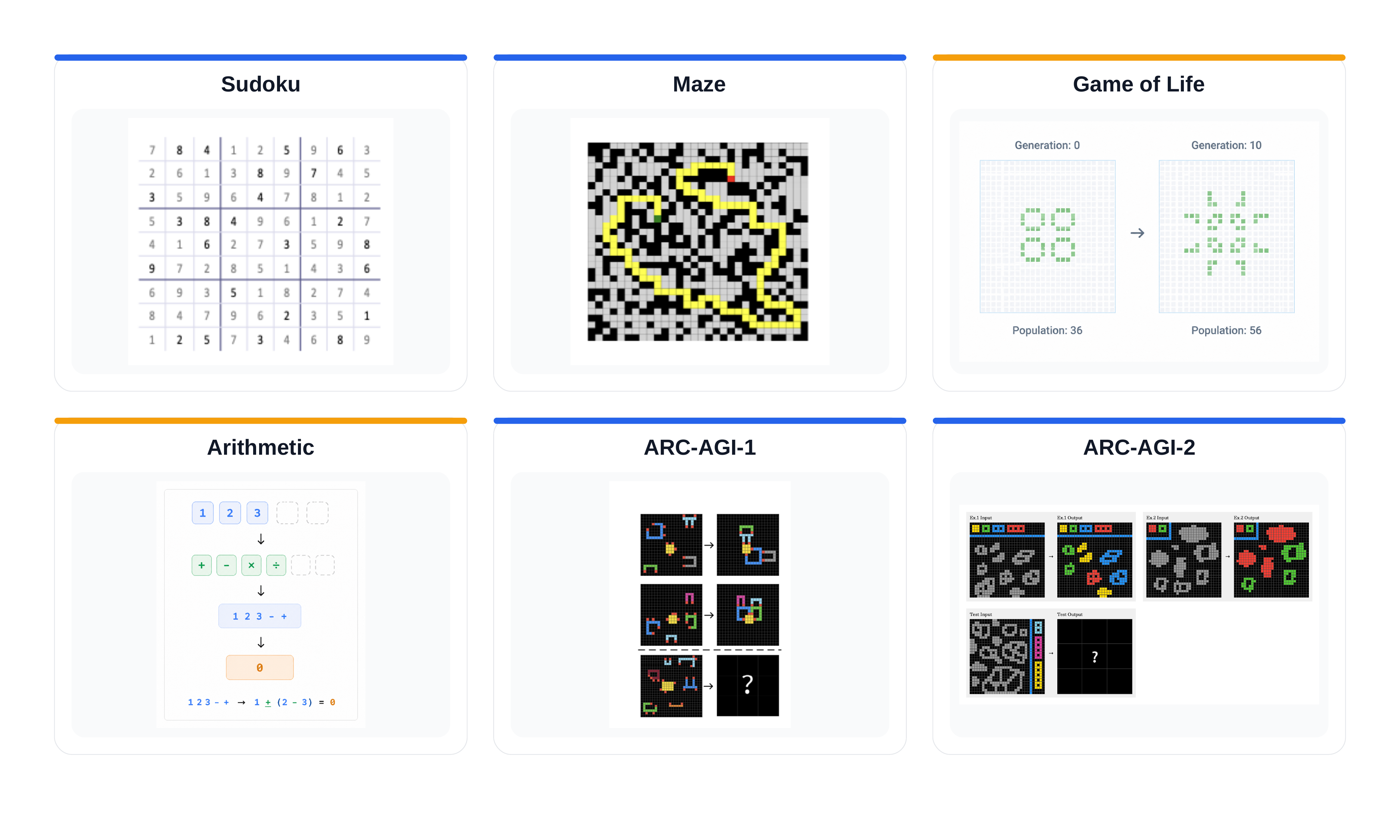}
    \caption{
        Evaluation domains used in our study.
        Sudoku, Maze, and ARC-AGI follow established recursive-reasoning benchmarks,
        while Game of Life and Arithmetic provide controlled out-of-distribution evaluation.
    }
    \label{fig:datasets}
\end{figure}

Prior work on recursive reasoning models evaluates on a small set of algorithmic
benchmarks, most prominently Sudoku, Maze, and ARC-AGI
\citep{wang2025hrm,jolicoeurmartineau2025trm,gao2025urm}.
We retain these benchmarks for direct comparability and add two domains,
Game of Life and Arithmetic, designed to provide explicit and controllable
out-of-distribution evaluation.
The resulting suite contains six domains spanning constraint satisfaction,
search, iterative dynamics, symbolic composition, and abstract rule induction
(Figure~\ref{fig:datasets}).

All tasks use the same interface: the input is represented as a padded token
sequence, and the model predicts the complete target without intermediate
supervision or chain-of-thought traces.

\FloatBarrier

\paragraph{Sudoku.}
We use the \texttt{sudoku-extreme} dataset following HRM
\citep{wang2025hrm}.
Each $9\times9$ puzzle is flattened into 81 tokens, with the completed grid as
the target. We train without symmetry-based augmentation.

\paragraph{Maze.}
We use the \texttt{maze-30x30-hard} dataset following HRM
\citep{wang2025hrm}.
Each $30\times30$ maze is flattened into 900 tokens, and the target is the same
grid with a shortest start-to-goal path marked.

\paragraph{Game of Life.}
We generate 200,000 Conway's Game of Life trajectories from random binary
grids of size up to $17\times17$.
Each example specifies an initial configuration and a step count $k$, and the
target is the configuration after exactly $k$ updates.
This makes the required number of sequential rule applications directly
controllable and enables evaluation of extrapolation beyond training horizons.

\paragraph{Arithmetic.}
We generate reverse-Polish expressions containing three to eight operands from
$\{1,\ldots,9\}$ and operators from $\{+,-,\times,\div\}$.
Operators are masked in the input while the final value is given, and the model
must recover the operator sequence.
This produces a constrained combinatorial search problem with up to $4^7$
candidate assignments.

\paragraph{ARC-AGI.}
We evaluate on ARC-AGI-1 and ARC-AGI-2 using the preprocessing and augmentation
protocol of prior recursive models
\citep{wang2025hrm,jolicoeurmartineau2025trm,gao2025urm}.
ARC-AGI-1 is combined with ConceptARC, while ARC-AGI-2 is kept separate.
We follow the same transductive training protocol used by these models;
full preprocessing and augmentation details are provided in
Appendix~\ref{app:data}.

\paragraph{Out-of-distribution evaluation.}
For Game of Life, we evaluate two forms of generalization.
First, $15\%$ of initial patterns are held out entirely.
Second, for patterns observed during training, we evaluate predictions
1, 2, 3, and 10 steps beyond their training horizon.

Arithmetic is split by operand multiset, so test expressions contain digit
combinations absent from training.
We additionally separate targets within the training value range, $[0,101]$,
from targets outside it, $[102,201]$.
Sudoku, Maze, and ARC use their standard evaluation splits.

\section{Stable Recursive Reasoning}
\label{sec:method}

Previous recursive reasoning models change multiple factors simultaneously,
including architecture, gradient propagation, and optimization procedure. We
therefore isolate these factors and construct a unified recipe based on
recurrent-state stabilization, credit assignment, and optimization stability.
The resulting model combines shared recurrent computation with controlled state
updates, intermediate gradient propagation, and recurrence-specific
regularization.

The final configuration is summarized in Section~\ref{sec:configuration}, while
the contribution of each component is evaluated through controlled ablations in
Section~\ref{sec:stabilization-ablation}.

\subsection{Recursive computation}

We use the nested recurrence formulation of HRM, where $z_L$ and $z_H$
represent low- and high-timescale recurrent states rather than separate
networks, and $x$ denotes input embeddings fixed within an Adaptive Computation
Time (ACT) step. Within each high-level cycle, the low-level state is updated
$L_{\mathrm{cycles}}$ times as
\[
z_L \leftarrow L(z_L + z_H + x),
\]
followed by the high-level update
\[
z_H \leftarrow H(z_H + z_L).
\]

The cycle is repeated $H_{\mathrm{cycles}}$ times. Unlike HRM, our default
configuration shares the parameters of the recurrent transformations;
Section~\ref{sec:hierarchy} evaluates whether explicit high/low-level parameter
separation provides additional benefits.

\subsection{Controlled recurrent-state updates}

Repeatedly applying the same transformation creates a unique optimization
challenge: small errors in recurrent updates accumulate over depth. Directly
replacing the recurrent state allows both the magnitude and the rate of
refinement to drift along long trajectories.

Motivated by this instability, we introduce a controlled refinement mechanism consisting of update bounding, learned gating, and post-cycle normalization. We first compute the
candidate update and bound its magnitude relative to the current state:
\[
\tilde{z}_L = L(z_L + z_H + x),
\qquad
\Delta_L = \tilde{z}_L-z_L,
\qquad
r = \frac{\|\Delta_L\|_2}{\|z_L\|_2+\epsilon},
\qquad
\hat{\Delta}_L =
\frac{\Delta_L}{\max(r/\tau,1)}.
\]

A learned gate then controls how much of the bounded update is applied:
\[
\alpha = \sigma\!\left(w^T(z_H+z_L+x)+b\right),
\qquad
z'_L =
\operatorname{Norm}\!\left(
z_L+\alpha\odot\hat{\Delta}_L
\right),
\]
where $\odot$ denotes element-wise multiplication, with $\alpha$ broadcast
over the hidden dimension.

Together, these mechanisms constrain the recurrent trajectory while preserving
the ability of different token positions to refine at different rates.

\subsection{Recurrence-consistent regularization}

Because recursive models repeatedly apply the same parameters along a single
latent trajectory, we reuse one dropout mask across all recurrent cycles and
ACT steps rather than sampling independent masks at each application. We also
add small norm-scaled Gaussian noise to the input embeddings and recurrent
states to discourage brittle trajectories. Exact mask placement and
regularization strengths are given in Appendix~\ref{app:training-details}.

\subsection{Training the recursion}

The model executes the full recurrent trajectory, while gradients propagate
only through the final $K_H$ high-level and $K_L$ low-level cycles. Unless
stated otherwise, we use $K_H=K_L=2$, following the controlled study in
Section~\ref{sec:gradient-path}. We optimize with Adam-atan2, global gradient
clipping at 1, and EMA weights for evaluation, and use large physical batches
rather than equivalent gradient accumulation following
Section~\ref{sec:physical-batch}.

\subsection{Final recursive reasoning recipe}
\label{sec:configuration}

The final model combines the design choices identified in our controlled study.
Rather than increasing architectural complexity, we focus on stabilizing the
dynamics of recurrent computation. The resulting configuration is used for all
main experiments in Section~\ref{sec:main-results}.

The model uses a shared Transformer block recurrently applied to latent states.
Each Adaptive Computation Time (ACT) step performs
$(H_{\mathrm{cycles}},L_{\mathrm{cycles}})=(4,2)$ recurrent updates. During
training, gradients are propagated through the final
$(K_L,K_H)=(2,2)$ recurrent updates of the trajectory.

The recurrent state is updated using bounded updates, learned gating, and
post-cycle normalization. Training uses recurrence-consistent dropout,
relative state and embedding noise, large physical batches, gradient clipping,
and exponential moving average (EMA) weights.

\begin{figure}[H]
    \centering
    \includegraphics[width=0.98\linewidth]{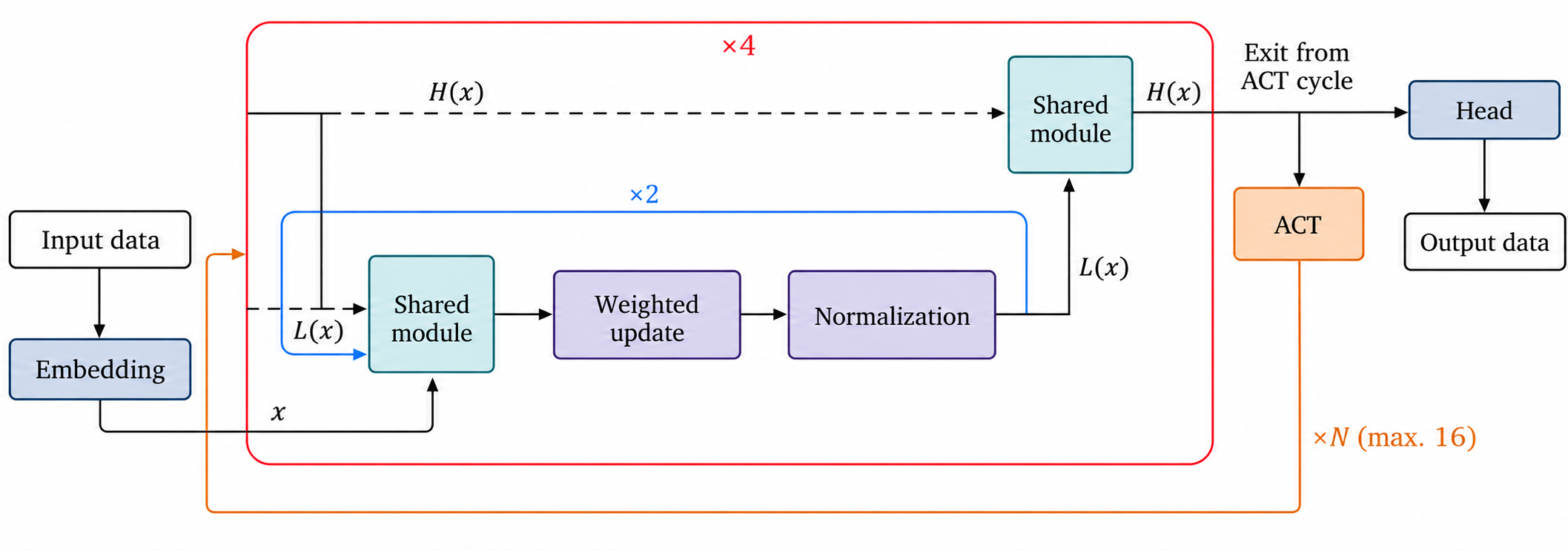}
    \caption{
        Overview of the final recursive reasoning architecture.
        A shared Transformer block repeatedly refines latent recurrent states.
        Controlled state updates and truncated gradient propagation stabilize
        the recursive trajectory.
    }
    \label{fig:architecture}
\end{figure}

The final architecture is summarized in Figure~\ref{fig:architecture}, and the
complete training recipe is given in Table~\ref{tab:recipe_comparison}.

\begin{table}[!ht]
\centering
\caption{Comparison of recursive reasoning recipes.}
\label{tab:recipe_comparison}

\setlength{\tabcolsep}{3.3pt}
\renewcommand{\arraystretch}{1.08}

\begin{tabular}{lcccc}
\toprule
\textbf{Setup}
& \textbf{HRM}
& \textbf{TRM}
& \textbf{URM}
& \textbf{Ours}
\\
\midrule

Parameters
& 27M
& 7M
& 7M
& 13.6M
\\

Modules
& 2 ($H+L$)
& 1 shared
& 1 shared
& 1 shared
\\

Layers
& $4+4$
& 2
& 2*
& 4
\\

Hidden size
& 512
& 512
& 512
& 512
\\

Cycles $(H,L)$
& $(2,2)$
& $(3,6)$
& $(3,6)$
& $(4,2)$
\\

Gradient path
& Last $H$ + last $L$
& Last $H$ + all $L$
& Last $H$ + all $L$
& $(K_L,K_H)=(2,2)$
\\

State stabilization
& --
& --
& --
& Bounded + gated + norm.
\\

Convolution
& No
& No
& Depthwise
& No**
\\

Physical batch
& 768
& 768
& 768
& Domain-specific; up to 8192
\\

Train stabilization
& Standard
& EMA
& EMA
& Consistent dropout + noise
\\

\bottomrule
\end{tabular}

\vspace{1mm}

\raggedright
\tiny
* For URM, we report the 2-layer configuration selected in our experiments.

** A convolutional variant is used only for ARC experiments and is not part
of the default recipe.
\end{table}
\label{sec:results}

\subsection{Evaluation and Main Results}
\label{sec:main-results}

We report exact accuracy on all domains and pass@1/pass@2 on ARC-AGI.
Adaptive halting is disabled during the main evaluation, and all examples
execute the full ACT budget. For test-time scaling, we extend this budget
beyond its training-time value.

We compare against HRM, TRM, and URM reconstructed in our codebase with matched data representation, tokenization, and evaluation. Each baseline retains its method-specific recurrent architecture and training recipe, while our model uses the recipe developed in \ref{sec:method}. Thus, this comparison evaluates complete recursive-reasoning recipes rather than isolating architecture alone; architecture-specific effects are studied separately in \ref{sec:ablations}. Full baseline configurations, ARC aggregation
details, and inference-cost accounting are provided in
Appendix~\ref{app:evaluation-details}.

\begin{table*}[!ht]
\centering
\caption{
Exact accuracy on the evaluation domains, in percent, from EMA weights.
Arithmetic is split into the in-distribution and out-of-distribution target
buckets of Section~\ref{sec:data}; Game of Life is reported on its in-range
set and as the mean over its out-of-distribution sets. ARC-AGI is evaluated
with pass@1 and pass@2. HRM, TRM, URM, and the dense control are retrained in
our codebase; general-purpose models are evaluated from released checkpoints
without fine-tuning. Best values are in bold.
}
\label{tab:main-results}
\resizebox{\textwidth}{!}{
\begin{tabular}{lccccccc}
\toprule
Metric
& Dense
& Qwen3-30B-A3B
& Qwen3.6-27B
& HRM
& TRM
& URM
& Ours \\
\midrule
Arithmetic ID
& 78.50 & 0.10 & 0.00 & 92.00 & \textbf{98.80} & 95.60 & 96.34 \\
Arithmetic OOD
& 7.49 & 0.00 & 0.00 & 13.80 & 36.20 & 28.50 & \textbf{71.16} \\
Game of Life ID
& 8.17 & 0.00 & 0.00 & 5.00 & 7.11 & 55.42 & \textbf{66.08} \\
Game of Life OOD
& 7.54 & 0.00 & 0.00 & 0.30 & 7.14 & 55.26 & \textbf{65.80} \\
Maze
& 8.00 & 0.20 & 0.00 & 71.10 & 80.00 & 80.20 & \textbf{84.70} \\
Sudoku
& 18.30 & 0.00 & 0.00 & 50.00 & 87.40 & 77.60 & \textbf{98.41} \\
ARC-AGI-1 pass@1
& 30.75 & 25.00 & 30.00 & 32.00 & 40.00 & 49.75 & \textbf{53.00} \\
ARC-AGI-1 pass@2
& 33.50 & 31.50 & 36.25 & 40.30 & 44.60 & 53.25 & \textbf{59.50} \\
ARC-AGI-2 pass@1
& 1.67 & 0.83 & 0.00 & 3.33 & 2.50 & 5.83 & \textbf{10.83} \\
ARC-AGI-2 pass@2
& 1.67 & 0.83 & 0.83 & 4.17 & 2.50 & 7.50 & \textbf{11.67} \\
\bottomrule
\end{tabular}
}
\end{table*}

Table~\ref{tab:main-results} summarizes the main comparison. Our model
achieves the strongest performance among the evaluated recursive baselines on
nearly all reported metrics. The largest improvement appears on Arithmetic
OOD, where accuracy increases from 36.20\% for the strongest recursive
baseline to 71.16\%.

The effect of recursion depends on the domain. On Sudoku and Maze, recursive
models substantially outperform the non-recurrent control. Game of Life
separates the recursive baselines: TRM remains close to the dense model,
whereas URM and our model achieve higher accuracy and retain similar
performance under distribution shift. ARC behaves differently: on ARC-AGI-1
the dense control is already competitive, while all systems remain limited on
ARC-AGI-2.

Overall, the largest gains appear on explicit generalization tests rather than
uniformly across all domains. Section~\ref{sec:ablations} analyzes which
optimization and architectural choices account for these differences.

\section{Controlled Ablations and Scaling Limits}
\label{sec:ablations}

We isolate the main architectural and optimization choices of
Section~\ref{sec:method} under controlled comparisons. Unless stated otherwise,
each experiment changes one factor while keeping the remaining training pipeline
fixed. We then test whether the resulting recipe continues to improve with more
model capacity, more tasks, or more inference-time computation.

\subsection{Gradient Path Length}
\label{sec:gradient-path}

We first isolate the effect of credit assignment through the recurrent
trajectory. The forward recursion is fixed at
$L_{\mathrm{cycles}}=2$ and $H_{\mathrm{cycles}}=4$, while only the backward
horizon is varied. Gradients propagate through the final $K_H$ high-level
cycles and, within each of them, through the final $K_L$ low-level cycles.
All eight runs use the same Arithmetic data, seed, optimizer, schedule,
architecture, and forward computation.

\begin{figure}[!htbp]
    \centering
    \includegraphics[
        width=\linewidth,
        height=0.22\textheight,
        keepaspectratio
    ]{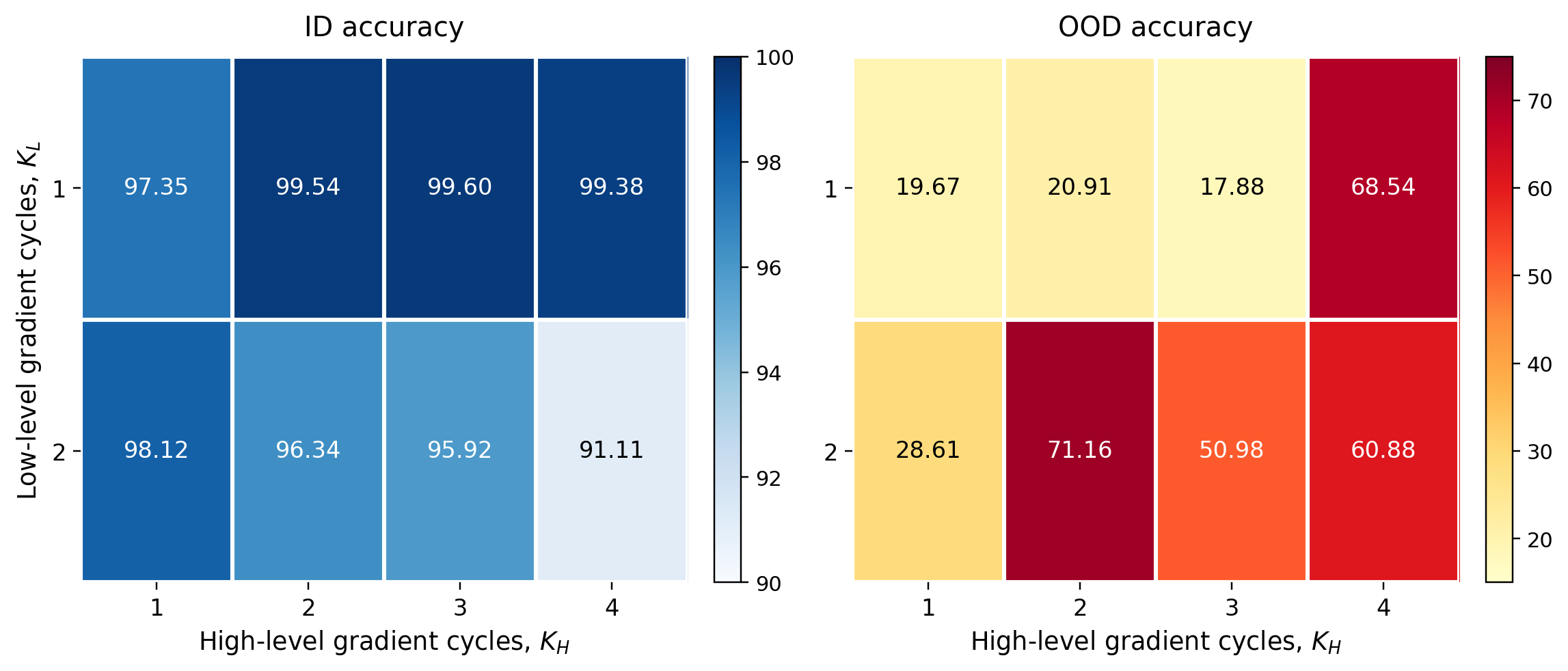}
    \caption{
        Exact accuracy on Arithmetic as a function of the differentiated
        gradient horizon. Forward recursion is fixed at
        $L_{\mathrm{cycles}}=2$ and $H_{\mathrm{cycles}}=4$; only
        $(K_L,K_H)$ varies. Left: in-distribution accuracy. Right:
        out-of-distribution accuracy.
    }
    \label{fig:gradient-path}
\end{figure}

Figure~\ref{fig:gradient-path} reveals a strongly non-monotonic relationship
between gradient horizon and generalization. The intermediate setting
$(K_L,K_H)=(2,2)$ achieves the highest OOD accuracy of 71.16\%, whereas
$(1,3)$ reaches the highest ID accuracy of 99.60\% but generalizes poorly,
obtaining only 17.88\% OOD.

Importantly, this difference cannot be explained by gradient-path length alone:
$(1,3)$ and $(2,2)$ differentiate the same number of recurrent applications but
produce sharply different OOD behavior. Thus, both the extent and placement of
gradient flow through the nested recurrence matter. On Arithmetic, the
intermediate $(2,2)$ setting gives the strongest out-of-distribution
generalization among the tested horizons while retaining high ID accuracy.

\FloatBarrier

\subsection{Physical Batch vs.\ Gradient Accumulation}
\label{sec:physical-batch}

We next test whether gradient accumulation reproduces the optimization behavior
of a genuinely larger physical batch. On Game of Life, we fix the effective
batch size at 8192 and vary the physical batch size together with the number of
accumulation steps, while keeping all other training settings fixed.

\begin{figure}[!htbp]
    \centering
    \includegraphics[
        width=0.90\linewidth,
        keepaspectratio
    ]{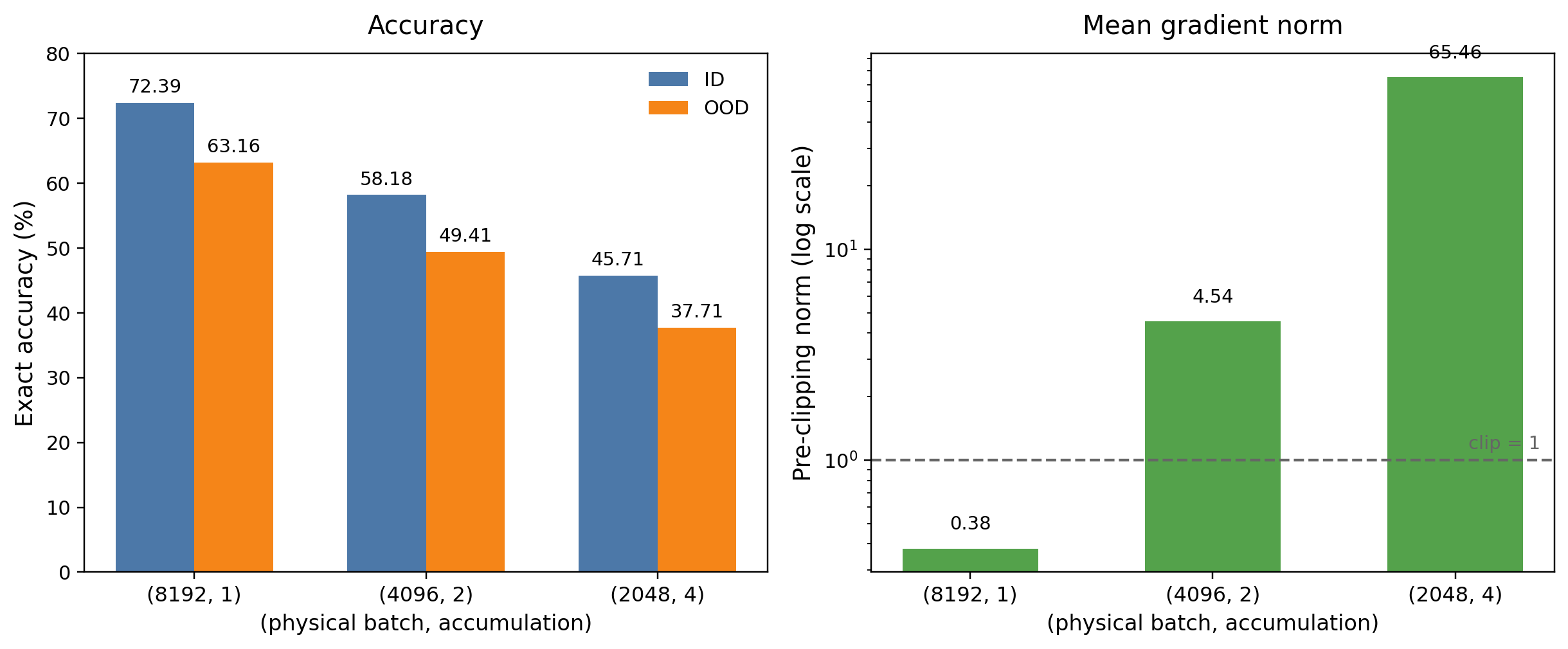}
    \caption{
        Effect of physical batch size versus gradient accumulation at a fixed
        effective batch size of 8192 on Game of Life. Configurations correspond
        to physical batch sizes of 8192, 4096, and 2048 with 1, 2, and 4
        accumulation steps, respectively. Left: ID and OOD accuracy.
        Right: mean pre-clipping gradient norm.
    }
    \label{fig:physical-batch}
\end{figure}

Figure~\ref{fig:physical-batch} shows that equal effective batch size does not
produce equivalent training dynamics. OOD accuracy decreases from 63.16\% with
a physical batch of 8192 to 37.71\% with four accumulation steps, while the mean
pre-clipping gradient norm increases from 0.38 to 65.46.

The effect is therefore not explained by the nominal number of examples
contributing to each optimizer update. In this setting, physical batch size is
an optimization parameter in its own right, and replacing it with gradient
accumulation substantially changes both stability and final accuracy. We
therefore report physical and effective batch sizes separately throughout the
study.

\subsection{Is the Hierarchy Necessary?}
\label{sec:hierarchy}

HRM introduces separate high- and low-level recurrent modules, whereas TRM and
URM share parameters across the two timescales. To isolate this architectural
choice, we compare shared and separate recurrent modules at matched parameter
count while keeping the remaining training configuration fixed.

Figure~\ref{fig:hierarchy-ablation} shows that neither design consistently
dominates across domains. Shared modules perform better on Maze and ARC-AGI-1,
while separate modules provide small gains on Sudoku and Game of Life; the two
configurations are nearly identical on Arithmetic.

These results indicate that explicit high/low-level parameter separation is not
a universal source of improvement. A simpler shared recurrent module remains
competitive across all studied domains and is preferable on several of them,
suggesting that the gains of recursive reasoning do not depend on an explicit
hierarchical decomposition.

\subsection{Stabilization Mechanisms}
\label{sec:stabilization-ablation}

We evaluate the stabilization recipe cumulatively on Game of Life and Sudoku,
focusing on both final accuracy and sensitivity to random seed. Starting from
the base configuration, we progressively add gating, gradient clipping,
recurrence-consistent dropout, and relative state and embedding noise.
Post-cycle normalization and EMA are kept fixed throughout.
Table~\ref{tab:stabilization} reports mean $\pm$ standard deviation over three
seeds.

\begin{table}[!htbp]
\centering
\caption{
Cumulative stabilization ablation on Game of Life and Sudoku.
Post-cycle normalization and EMA are fixed in all configurations.
Values report mean $\pm$ standard deviation over three seeds.
Best values are in bold and second-best values are underlined.
}
\label{tab:stabilization}
\small
\setlength{\tabcolsep}{5pt}
\renewcommand{\arraystretch}{1.12}

\begin{tabular}{lccc}
\toprule
\textbf{Configuration}
& \textbf{GoL ID}
& \textbf{GoL OOD}
& \textbf{Sudoku}
\\
\midrule

Base
& $61.95 \pm 3.32$
& $63.87 \pm 2.90$
& $97.29 \pm 1.06$
\\

+ gate
& $\underline{64.24 \pm 2.59}$
& $\mathbf{65.89 \pm 2.77}$
& $98.41 \pm 0.25$
\\

+ grad clip
& $62.41 \pm 1.56$
& $63.87 \pm 1.54$
& $\mathbf{98.50 \pm 0.20}$
\\

+ dropout
& $61.82 \pm 0.34$
& $62.09 \pm 0.41$
& $98.36 \pm 0.25$
\\

+ relative noise (final)
& $\mathbf{66.08 \pm 0.11}$
& $\underline{65.80 \pm 0.13}$
& $\underline{98.41 \pm 0.21}$
\\

\bottomrule
\end{tabular}
\end{table}

The stabilization recipe has a pronounced effect on training reliability,
especially on the more challenging Game of Life domain. The base configuration
is highly seed-sensitive, with standard deviations of 3.32 and 2.90 points on
the ID and OOD splits. With the complete recipe, these decrease to only 0.11
and 0.13 points, while accuracy rises to 66.08\% ID and 65.80\% OOD.

Sudoku shows the same pattern in a higher-accuracy regime: variability falls
from 1.06 points in the base configuration to approximately 0.2 points in the
stabilized variants while mean accuracy remains near 98\%. The main effect of
the recipe is therefore not merely to preserve performance, but to turn a
seed-sensitive recurrent optimization problem into a substantially more
reproducible one. The benefit is strongest on the harder domain, where the
final configuration combines both high accuracy and very low run-to-run
variance.

\FloatBarrier

\subsection{Limits of Scaling Recursive Computation}
\label{sec:scaling}

Finally, we test whether the resulting recipe continues to improve with
additional model capacity or inference-time computation. Increasing the
recurrent block from four to eight layers, increasing hidden size, or adding
attention heads does not yield consistent accuracy gains across the tested
settings. We therefore retain the 13.6M-parameter configuration as the default
model rather than increasing capacity further.

\begin{figure*}[!t]
    \centering

    \begin{subfigure}[t]{0.48\textwidth}
        \centering
        \includegraphics[
            width=\linewidth,
            height=0.22\textheight,
            keepaspectratio
        ]{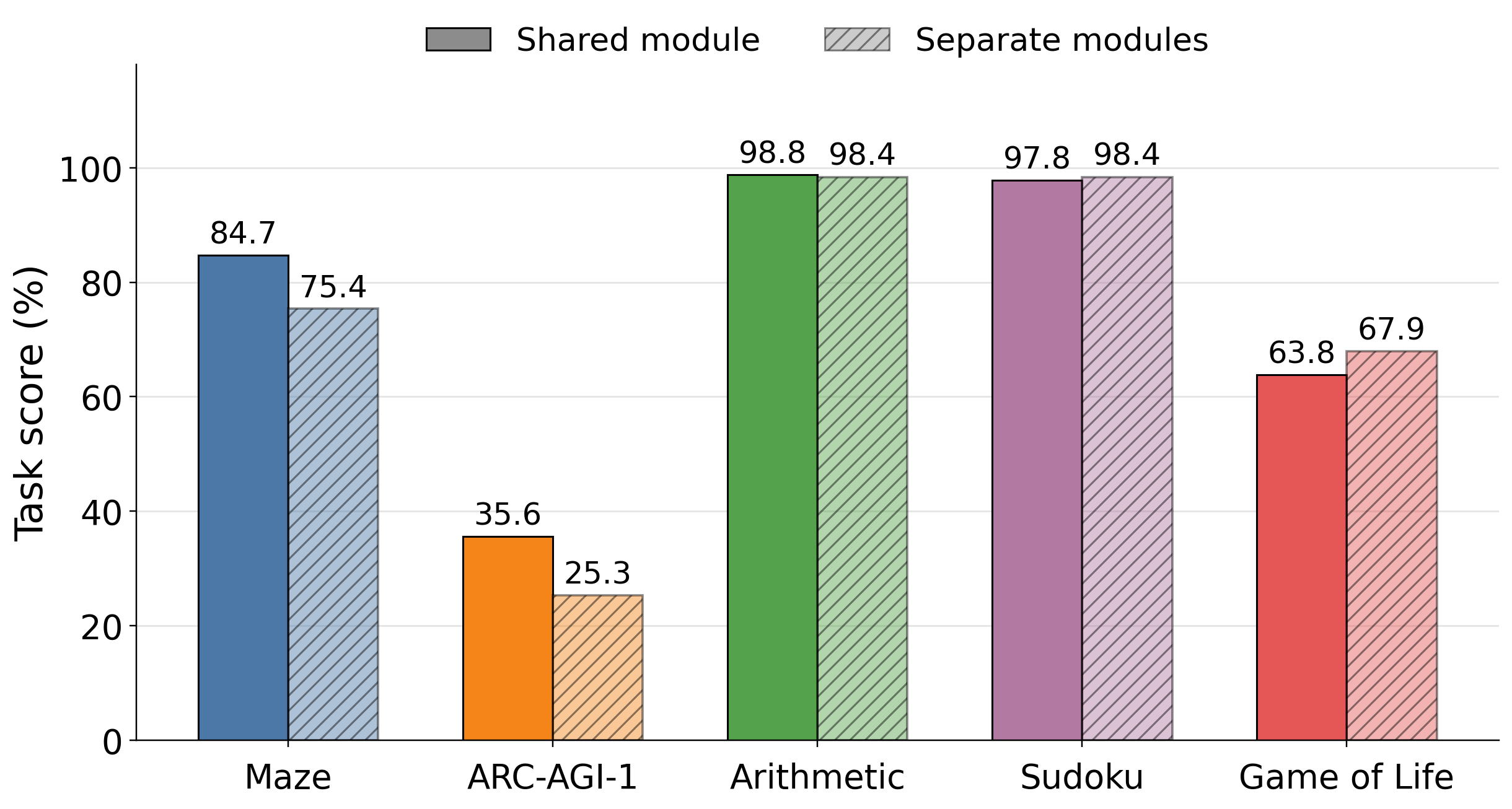}
        \caption{
            Effect of shared versus separate recurrent modules at matched
            parameter count.
        }
        \label{fig:hierarchy-ablation}
    \end{subfigure}
    \hfill
    \begin{subfigure}[t]{0.48\textwidth}
        \centering
        \includegraphics[
            width=\linewidth,
            height=0.22\textheight,
            keepaspectratio
        ]{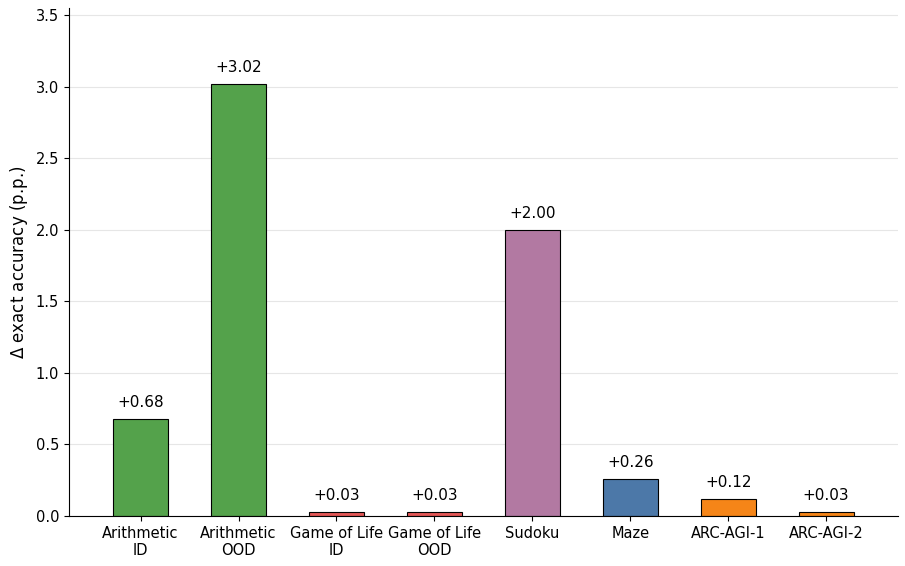}
        \caption{
            Accuracy change from increasing the ACT budget beyond the
            training-time value.
        }
        \label{fig:test-time-scaling}
    \end{subfigure}

    \caption{
        Architectural and test-time scaling ablations.
        (a) Effect of sharing parameters between recurrent timescales.
        (b) Effect of additional recurrent computation at inference time.
    }
    \label{fig:combined-ablations}
\end{figure*}

Test-time computation exhibits a similarly task-dependent pattern.
Figure~\ref{fig:test-time-scaling} shows that increasing the ACT budget provides
the largest gains on Arithmetic and Sudoku, while improvements on Game of Life,
Maze, and ARC-AGI remain limited.

Together, these results show that recursive reasoning does not improve simply by
adding more parameters or more recurrent steps. Capacity and test-time compute
help selectively, while the largest and most consistent improvements in our
study come from how the recurrent computation is optimized and stabilized.

\section{Conclusion}
\label{sec:conclusion}

We presented a controlled study of recursive reasoning models, separating
architectural and optimization choices that are coupled in previous systems.
Across six algorithmic domains, we find that stable recursive computation
depends strongly on how recurrence is trained and stabilized.

Our experiments identify three important factors. First, gradient propagation
through the recurrent trajectory has an intermediate optimum: neither aggressive
truncation nor full backpropagation provides the best generalization. Second,
physical batch size affects optimization beyond effective batch size, with
large physical batches substantially improving stability and accuracy compared
with gradient accumulation. Third, controlling recurrent-state updates through
bounded, gated, and normalized transformations enables reliable training.

At the same time, adding capacity does not guarantee better reasoning.
Explicit hierarchy, larger recurrent blocks, and additional test-time
computation provide domain-dependent rather than universal gains. These results
suggest that recursive reasoning should be viewed as an optimization problem as
much as an architectural one: the effectiveness of recurrence depends on how
information and gradients are propagated through the recurrent trajectory.

\subsection*{AI use statement}

We used generative AI tools for manuscript preparation, literature discovery, and code
refactoring. Generative AI was used to draft, revise, and polish parts of the manuscript and to
assist with retrieval and discovery of relevant prior work. The authors independently verified all
citations against the original sources and reviewed all AI-assisted text. The research questions,
experimental design, experiments, measurements, analysis, and conclusions are the work of the
authors.

Generative AI was also used interactively to assist with refactoring existing research code.
AI-suggested changes were manually reviewed, adapted where necessary, integrated into the
repository, and tested by the authors. The experimental configurations and experiments reported
in the paper were controlled and executed by the authors. We reviewed all AI-assisted material
and take full responsibility for the content of this paper, including its text, code, claims, and
reported results.

\subsection*{Reproducibility statement}

The submission includes the full source code used for the experiments, together with all reported
experiment configurations and the scripts used to generate the Arithmetic and Game of Life
datasets. Reported experiments can be reproduced from their corresponding configuration files,
which specify the model, optimization, data, evaluation, and random-seed settings. Random seeds
are propagated to Python, NumPy, PyTorch, and data-loader workers.

Section~\ref{sec:data} describes the evaluation domains and distribution shifts. The model and
training procedure are described in Section~\ref{sec:method} and Appendix~\ref{app:training-details}, and the evaluation protocol and
baseline details are given in Section~\ref{sec:main-results} and Appendix~\ref{app:evaluation-details}. In particular, adaptive halting is
disabled for the main evaluation and all examples execute the full evaluation budget.

Where results are aggregated over multiple seeds, this is stated explicitly in the corresponding
table or figure caption; results obtained from a single run are reported as such.

\bibliography{iclr2027_conference}

\begin{thebibliography}{16}
\providecommand{\natexlab}[1]{#1}
\providecommand{\url}[1]{\texttt{#1}}
\expandafter\ifx\csname urlstyle\endcsname\relax
  \providecommand{\doi}[1]{doi: #1}\else
  \providecommand{\doi}{doi: \begingroup \urlstyle{rm}\Url}\fi

\bibitem[Bae et~al.(2025{\natexlab{a}})Bae, Fisch, Harutyunyan, Ji, Kim, and Schuster]{bae2025relaxed}
Sangmin Bae, Adam Fisch, Hrayr Harutyunyan, Ziwei Ji, Seungyeon Kim, and Tal Schuster.
\newblock Relaxed recursive transformers: Effective parameter sharing with layer-wise {LoRA}.
\newblock In \emph{The Thirteenth International Conference on Learning Representations}, 2025{\natexlab{a}}.
\newblock URL \url{https://openreview.net/forum?id=WwpYSOkkCt}.

\bibitem[Bae et~al.(2025{\natexlab{b}})Bae, Kim, Bayat, Kim, Ha, Schuster, Fisch, Harutyunyan, Ji, Courville, and Yun]{bae2025mixture}
Sangmin Bae, Yujin Kim, Reza Bayat, Sungnyun Kim, Jiyoun Ha, Tal Schuster, Adam Fisch, Hrayr Harutyunyan, Ziwei Ji, Aaron Courville, and Se-Young Yun.
\newblock Mixture-of-recursions: Learning dynamic recursive depths for adaptive token-level computation.
\newblock In \emph{Advances in Neural Information Processing Systems}, 2025{\natexlab{b}}.
\newblock URL \url{https://arxiv.org/abs/2507.10524}.

\bibitem[Dehghani et~al.(2019)Dehghani, Gouws, Vinyals, Uszkoreit, and Kaiser]{dehghani2019universal}
Mostafa Dehghani, Stephan Gouws, Oriol Vinyals, Jakob Uszkoreit, and {\L}ukasz Kaiser.
\newblock Universal transformers.
\newblock In \emph{Proceedings of the 7th International Conference on Learning Representations}, 2019.
\newblock URL \url{https://arxiv.org/abs/1807.03819}.

\bibitem[Fan et~al.(2025)Fan, Du, Ramchandran, and Lee]{fan2025looped}
Ying Fan, Yilun Du, Kannan Ramchandran, and Kangwook Lee.
\newblock Looped transformers for length generalization.
\newblock In \emph{Proceedings of the 13th International Conference on Learning Representations}, 2025.
\newblock URL \url{https://arxiv.org/abs/2409.15647}.

\bibitem[Fu et~al.(2025)Fu, You, Chen, Dai, Yang, and Wang]{fu2025thinkhard}
Tianyu Fu, Yichen You, Zekai Chen, Guohao Dai, Huazhong Yang, and Yu~Wang.
\newblock Think-at-hard: Dynamic looped transformers for improved reasoning, 2025.
\newblock URL \url{https://arxiv.org/abs/2511.08577}.

\bibitem[Gao et~al.(2026)Gao, Chen, Xiao, Yang, Tao, Zhou, and Dai]{gao2026loopies}
Zitian Gao, Yilong Chen, Yihao Xiao, Xinyu Yang, Ran Tao, Joey Zhou, and Bryan Dai.
\newblock Loop the loopies!, 2026.
\newblock URL \url{https://arxiv.org/abs/2607.16051}.

\bibitem[Geiping et~al.(2025)Geiping, McLeish, Jain, Kirchenbauer, Singh, Bartoldson, Kailkhura, Bhatele, and Goldstein]{geiping2025scaling}
Jonas Geiping, Sean McLeish, Neel Jain, John Kirchenbauer, Siddharth Singh, Brian~R. Bartoldson, Bhavya Kailkhura, Abhinav Bhatele, and Tom Goldstein.
\newblock Scaling up test-time compute with latent reasoning: A recurrent depth approach.
\newblock In \emph{Advances in Neural Information Processing Systems}, 2025.
\newblock URL \url{https://arxiv.org/abs/2502.05171}.

\bibitem[Graves(2016)]{graves2016act}
Alex Graves.
\newblock Adaptive computation time for recurrent neural networks.
\newblock \emph{arXiv preprint arXiv:1603.08983}, 2016.
\newblock URL \url{https://arxiv.org/abs/1603.08983}.

\bibitem[Movahedi et~al.(2026)Movahedi, Milovanovi{\'c}, Feigin, Theus, Hofmann, Boeva, Rusch, and Orvieto]{movahedi2026fixedpoint}
Sajad Movahedi, Vera Milovanovi{\'c}, Shlomo~Libo Feigin, Alexander Theus, Thomas Hofmann, Valentina Boeva, T.~Konstantin Rusch, and Antonio Orvieto.
\newblock Fixed-point reasoners: Stable and adaptive deep looped transformers, 2026.
\newblock URL \url{https://arxiv.org/abs/2606.18206}.
\newblock ICML 2026 Workshop.

\bibitem[Saunshi et~al.(2025)Saunshi, Dikkala, Li, Kumar, and Reddi]{saunshi2025latent}
Nikunj Saunshi, Nishanth Dikkala, Zhiyuan Li, Sanjiv Kumar, and Sashank~J. Reddi.
\newblock Reasoning with latent thoughts: On the power of looped transformers.
\newblock In \emph{Proceedings of the 13th International Conference on Learning Representations}, 2025.
\newblock URL \url{https://openreview.net/forum?id=Pr8o5llJ1O}.

\bibitem[Shao et~al.(2024)Shao, Wang, Zhu, Xu, Song, Bi, Zhang, Zhang, Li, Wu, and Guo]{shao2024deepseekmath}
Zhihong Shao, Peiyi Wang, Qihao Zhu, Runxin Xu, Junxiao Song, Xiao Bi, Haowei Zhang, Mingchuan Zhang, Y.~K. Li, Y.~Wu, and Daya Guo.
\newblock {DeepSeekMath}: Pushing the limits of mathematical reasoning in open language models, 2024.
\newblock URL \url{https://arxiv.org/abs/2402.03300}.

\bibitem[Wang et~al.(2026)Wang, Liu, Wang, Zhou, Sun, Wu, Zhen, Scimeca, and Yadkori]{wang2026hrmtext}
Guan Wang, Changling Liu, Chenyu Wang, Cai Zhou, Yuhao Sun, Yifei Wu, Shuai Zhen, Luca Scimeca, and Yasin~Abbasi Yadkori.
\newblock Hrm-text: Efficient pretraining beyond scaling, 2026.
\newblock URL \url{https://arxiv.org/abs/2605.20613}.

\bibitem[Williams \& Tureci(2026)Williams and Tureci]{williams2026rltt}
Jonathan Williams and Esin Tureci.
\newblock Prioritize the process, not just the outcome: Rewarding latent thought trajectories improves reasoning in looped language models.
\newblock In \emph{Proceedings of the 43rd International Conference on Machine Learning}, 2026.
\newblock URL \url{https://arxiv.org/abs/2602.10520}.

\bibitem[Yang et~al.(2025)Yang, Li, Yang, Zhang, Hui, Zheng, Yu, Gao, Huang, Lv, Zheng, Liu, Zhou, Huang, Hu, Ge, Wei, Lin, Tang, Yang, Tu, Zhang, Yang, Yang, Zhou, Zhou, Lin, Dang, Bao, Yang, Yu, Deng, Li, Xue, Li, Zhang, Wang, Zhu, Men, Liu, Luo, Li, Tang, Yin, Ren, Wang, Zhang, Ren, Fan, Su, Zhang, Zhang, Wan, Liu, Wang, Cui, Zhang, Zhou, and Qiu]{yang2025qwen3}
An~Yang, Anfeng Li, Baosong Yang, Beichen Zhang, Binyuan Hui, Bo~Zheng, Bowen Yu, Chang Gao, Chengen Huang, Chenxu Lv, Chujie Zheng, Dayiheng Liu, Fan Zhou, Fei Huang, Feng Hu, Hao Ge, Haoran Wei, Huan Lin, Jialong Tang, Jian Yang, Jianhong Tu, Jianwei Zhang, Jianxin Yang, Jiaxi Yang, Jing Zhou, Jingren Zhou, Junyang Lin, Kai Dang, Keqin Bao, Kexin Yang, Le~Yu, Lianghao Deng, Mei Li, Mingfeng Xue, Mingze Li, Pei Zhang, Peng Wang, Qin Zhu, Rui Men, Ruize Liu, Shuang Luo, Tianhao Li, Tianyi Tang, Wenbiao Yin, Xingzhang Ren, Xinyu Wang, Xinyu Zhang, Xuancheng Ren, Yang Fan, Yang Su, Yichang Zhang, Yinger Zhang, Yu~Wan, Yuqiong Liu, Zekun Wang, Zeyu Cui, Zhenru Zhang, Zhipeng Zhou, and Zihan Qiu.
\newblock {Qwen3} technical report.
\newblock \emph{arXiv preprint arXiv:2505.09388}, 2025.
\newblock URL \url{https://arxiv.org/abs/2505.09388}.

\bibitem[Yu et~al.(2025)]{yu2025dapo}
Qiying Yu et~al.
\newblock {DAPO}: An open-source {LLM} reinforcement learning system at scale, 2025.
\newblock URL \url{https://arxiv.org/abs/2503.14476}.

\bibitem[Zhu et~al.(2025)Zhu, Wang, Hua, Zhang, Li, Que, Wei, Wen, Yin, Xing, Li, Shi, Ma, Li, Kergan, Smith, Qu, Hui, Wu, Min, Huang, Zhou, Ye, Liu, Yang, Shi, Lin, Zhao, Cai, Zhang, Huang, Bengio, and Eshraghian]{zhu2025ouro}
Rui-Jie Zhu, Zixuan Wang, Kai Hua, Tianyu Zhang, Ziniu Li, Haoran Que, Boyi Wei, Zixin Wen, Fan Yin, He~Xing, Lu~Li, Jiajun Shi, Kaijing Ma, Shanda Li, Taylor Kergan, Andrew Smith, Xingwei Qu, Mude Hui, Bohong Wu, Qiyang Min, Hongzhi Huang, Xun Zhou, Wei Ye, Jiaheng Liu, Jian Yang, Yunfeng Shi, Chenghua Lin, Enduo Zhao, Tianle Cai, Ge~Zhang, Wenhao Huang, Yoshua Bengio, and Jason Eshraghian.
\newblock Scaling latent reasoning via looped language models.
\newblock \emph{arXiv preprint arXiv:2510.25741}, 2025.
\newblock URL \url{https://arxiv.org/abs/2510.25741}.

\end{thebibliography}


\begin{thebibliography}{14}
\providecommand{\natexlab}[1]{#1}
\providecommand{\url}[1]{\texttt{#1}}
\expandafter\ifx\csname urlstyle\endcsname\relax
  \providecommand{\doi}[1]{doi: #1}\else
  \providecommand{\doi}{doi: \begingroup \urlstyle{rm}\Url}\fi

\bibitem[Asadulaev et~al.(2026)Asadulaev, Banerjee, Karray, and Takac]{asadulaev2026latent}
Arip Asadulaev, Rayan Banerjee, Fakhri Karray, and Martin Takac.
\newblock Latent reasoning in trms is secretly a policy improvement operator, 2026.

\bibitem[Bae et~al.(2025)Bae, Kim, Bayat, Kim, Ha, Schuster, Fisch, Harutyunyan, Ji, Courville, and Yun]{bae2025mixture}
Sangmin Bae, Yujin Kim, Reza Bayat, Sungnyun Kim, Jiyoun Ha, Tal Schuster, Adam Fisch, Hrayr Harutyunyan, Ziwei Ji, Aaron Courville, and Se-Young Yun.
\newblock Mixture-of-recursions: Learning dynamic recursive depths for adaptive token-level computation.
\newblock In \emph{Advances in Neural Information Processing Systems}, 2025.
\newblock URL \url{https://arxiv.org/abs/2507.10524}.

\bibitem[Chollet(2019)]{chollet2019measure}
Fran\c{c}ois Chollet.
\newblock On the measure of intelligence.
\newblock \emph{arXiv preprint arXiv:1911.01547}, 2019.
\newblock URL \url{https://arxiv.org/abs/1911.01547}.

\bibitem[Chollet et~al.(2025)]{chollet2025arcagi2}
Fran\c{c}ois Chollet et~al.
\newblock {ARC-AGI-2}: A new challenge for frontier {AI} reasoning systems.
\newblock \emph{arXiv preprint arXiv:2505.11831}, 2025.
\newblock URL \url{https://arxiv.org/abs/2505.11831}.

\bibitem[Fan et~al.(2025)Fan, Du, Ramchandran, and Lee]{fan2025looped}
Ying Fan, Yilun Du, Kannan Ramchandran, and Kangwook Lee.
\newblock Looped transformers for length generalization.
\newblock In \emph{Proceedings of the 13th International Conference on Learning Representations}, 2025.
\newblock URL \url{https://arxiv.org/abs/2409.15647}.

\bibitem[Gao et~al.(2025)Gao, Chen, Xiao, Xing, Tao, Luo, Zhou, and Dai]{gao2025urm}
Zitian Gao, Lynx Chen, Yihao Xiao, He~Xing, Ran Tao, Haoming Luo, Joey Zhou, and Bryan Dai.
\newblock Universal reasoning model, 2025.
\newblock URL \url{https://arxiv.org/abs/2512.14693}.

\bibitem[Ge et~al.(2025)Ge, Liao, and Poggio]{ge2025perspectives}
Renee Ge, Qianli Liao, and Tomaso Poggio.
\newblock Hierarchical reasoning models: Perspectives and misconceptions, 2025.
\newblock URL \url{https://arxiv.org/abs/2510.00355}.

\bibitem[Geiping et~al.(2025)Geiping, McLeish, Jain, Kirchenbauer, Singh, Bartoldson, Kailkhura, Bhatele, and Goldstein]{geiping2025scaling}
Jonas Geiping, Sean McLeish, Neel Jain, John Kirchenbauer, Siddharth Singh, Brian~R. Bartoldson, Bhavya Kailkhura, Abhinav Bhatele, and Tom Goldstein.
\newblock Scaling up test-time compute with latent reasoning: A recurrent depth approach.
\newblock In \emph{Advances in Neural Information Processing Systems}, 2025.
\newblock URL \url{https://arxiv.org/abs/2502.05171}.

\bibitem[Jolicoeur-Martineau(2025)]{jolicoeurmartineau2025trm}
Alexia Jolicoeur-Martineau.
\newblock Less is more: Recursive reasoning with tiny networks, 2025.
\newblock URL \url{https://arxiv.org/abs/2510.04871}.

\bibitem[Macfarlane \& Bonnet(2025)Macfarlane and Bonnet]{macfarlane2025searching}
Matthew~V. Macfarlane and Clément Bonnet.
\newblock Searching latent program spaces, 2025.

\bibitem[Moskvichev et~al.(2023)Moskvichev, Odouard, and Mitchell]{moskvichev2023conceptarc}
Arseny Moskvichev, Victor~Vikram Odouard, and Melanie Mitchell.
\newblock The {ConceptARC} benchmark: Evaluating understanding and generalization in the {ARC} domain.
\newblock \emph{Transactions on Machine Learning Research}, 2023.
\newblock URL \url{https://arxiv.org/abs/2305.07141}.

\bibitem[Movahedi et~al.(2026)Movahedi, Milovanovi{\'c}, Feigin, Theus, Hofmann, Boeva, Rusch, and Orvieto]{movahedi2026fixedpoint}
Sajad Movahedi, Vera Milovanovi{\'c}, Shlomo~Libo Feigin, Alexander Theus, Thomas Hofmann, Valentina Boeva, T.~Konstantin Rusch, and Antonio Orvieto.
\newblock Fixed-point reasoners: Stable and adaptive deep looped transformers, 2026.
\newblock URL \url{https://arxiv.org/abs/2606.18206}.
\newblock ICML 2026 Workshop.

\bibitem[Saunshi et~al.(2025)Saunshi, Dikkala, Li, Kumar, and Reddi]{saunshi2025latent}
Nikunj Saunshi, Nishanth Dikkala, Zhiyuan Li, Sanjiv Kumar, and Sashank~J. Reddi.
\newblock Reasoning with latent thoughts: On the power of looped transformers.
\newblock In \emph{Proceedings of the 13th International Conference on Learning Representations}, 2025.
\newblock URL \url{https://openreview.net/forum?id=Pr8o5llJ1O}.

\bibitem[Wang et~al.(2025)Wang, Li, Sun, Chen, Liu, Wu, Lu, Song, and Abbasi-Yadkori]{wang2025hrm}
Guan Wang, Jin Li, Yuhao Sun, Xing Chen, Changling Liu, Yue Wu, Meng Lu, Sen Song, and Yasin Abbasi-Yadkori.
\newblock Hierarchical reasoning model, 2025.
\newblock URL \url{https://arxiv.org/abs/2506.21734}.

\end{thebibliography}
\bibliographystyle{iclr2027_conference}
\clearpage
\appendix
\section{Dataset Details}
\label{app:data}

\subsection{Common representation}

All domains are mapped to a common sequence prediction interface.
Inputs are represented as fixed-length padded token sequences, and the model predicts
the complete target without intermediate supervision or chain-of-thought traces.
For grid-based tasks, two-dimensional structures are flattened into token sequences.
Task-specific encoding details are given below.

\subsection{Sudoku}

We use the \texttt{sudoku-extreme} corpus following HRM
\citep{wang2025hrm}.
Each instance is a $9\times9$ Sudoku board, with empty cells represented by zeros.
The input board is flattened into a sequence of 81 tokens, and the target is the
corresponding completed board.

We train on the full training split without symmetry-based augmentation.
Thus, evaluation is performed on held-out puzzles rather than on transformed versions
of a smaller underlying puzzle set.

\subsection{Maze}

We use the \texttt{maze-30x30-hard} corpus following HRM
\citep{wang2025hrm}.
Each instance is a $30\times30$ grid over the symbols
\{\texttt{wall}, \texttt{free}, \texttt{start}, \texttt{goal}, \texttt{path}\}.
The input contains the maze together with the start and goal positions.
The target is the same grid with a shortest start-to-goal path marked.

Both the input and target grids are flattened into sequences of 900 tokens.

\subsection{Game of Life}

\paragraph{Generation.}
We generate 200,000 binary initial configurations.
Grid width and height are sampled independently and uniformly from $[2,17]$,
and each cell is initially alive with probability $0.4$.

Each configuration is evolved according to Conway's Game of Life using the
standard B3/S23 transition rule.
Evolution continues until the configuration becomes empty, enters a periodic orbit,
or reaches 50 evolution steps.
Patterns whose serialized representation exceeds 250 characters are discarded,
resulting in encoded sequences of at most 253 tokens.

Each example consists of an initial configuration and a requested evolution step $k$,
separated by a dedicated token.
The target is the configuration obtained after exactly $k$ applications of the
Game-of-Life transition rule.

This formulation makes the required number of sequential rule applications explicit:
increasing $k$ increases the length of the computation required to obtain the target
without changing the underlying transition rule.

\paragraph{Distribution shifts.}
We construct evaluation splits along two independent axes: initial configuration
and computational horizon.

For pattern generalization, $15\%$ of initial configurations are held out entirely
from training.
Examples derived from these configurations form the \emph{unseen-pattern} test set.

For horizon generalization, early evolution steps from the remaining configurations
are used for training.
Held-out steps from the same configurations that remain within the corresponding
training horizon form the \emph{in-range} evaluation split.
We additionally construct four extrapolation splits whose targets lie
1, 2, 3, and 10 evolution steps beyond the maximum horizon observed during training
for that configuration.

The resulting splits therefore separate generalization to unseen initial states
from extrapolation in the number of required recursive computations.
Each evaluation split is subsampled to 10,000 examples.

\subsection{Arithmetic}

\paragraph{Generation.}
We generate expressions in reverse Polish notation containing between three and
eight operands.
Operands are sampled from $\{1,\ldots,9\}$ and operators from
$\{+,-,\times,\div\}$.
Division is permitted only when it produces an exact integer result.

The resulting sequences contain at most 19 tokens.
At input time, every operator is replaced by a special \texttt{?} token and the
final value of the expression is appended to the input.
The model is trained to reconstruct the hidden operator sequence, with the loss
applied only at the masked operator positions.

For an expression containing $m$ operators, the unconstrained search space contains
up to $4^m$ possible operator assignments.
Since the longest generated expressions contain seven operators, the largest
search space contains $4^7$ assignments, constrained by the observed final value.

\paragraph{Distribution shifts.}
We construct train and test sets by splitting over operand multisets.
The split is $80/20$, and every test expression contains a multiset of operands that
does not occur in training.

Test examples are additionally partitioned according to the final expression value.
Targets in $[0,101]$ form the in-distribution evaluation set.
Targets in $[102,201]$ lie outside the range used for training and form the
out-of-distribution value split.

This construction therefore combines two forms of generalization:
unseen operand combinations and extrapolation to unseen target values.

\subsection{ARC-AGI}

We evaluate on ARC-AGI-1 \citep{chollet2019measure} and ARC-AGI-2
\citep{chollet2025arcagi2}.
ARC-AGI-1 is combined with ConceptARC \citep{moskvichev2023conceptarc},
while ARC-AGI-2 is kept separate.

We reuse the HRM data pipeline \citep{wang2025hrm}, including its encoding and
augmentation procedure.
Each ARC grid is embedded into a $30\times30$ canvas with padding and explicit
end-of-grid markers and is then flattened into a sequence of 900 tokens.
The vocabulary contains ten color tokens and two special symbols.

\paragraph{Augmentation.}
Each puzzle is expanded into 1000 augmented examples.
Augmentations compose the eight dihedral transformations of the grid with a random
permutation of the color vocabulary.
Training examples additionally receive a random translation within the
$30\times30$ canvas.

At evaluation time, predictions are mapped back to the original coordinate and
color systems by applying the inverse color permutation and inverse dihedral
transformation and are cropped according to the end-of-grid markers.

\paragraph{Transductive protocol.}
Following prior recursive-model evaluations, few-shot demonstrations are not provided
in context at inference time.
Instead, each demonstration pair belonging to a task is converted into a separate
supervised training example.

This also applies to tasks belonging to the evaluation split.
The held-out test input itself is not used for training, but demonstration pairs from
the same task are observed during training.
A learned task-specific puzzle embedding is the only channel connecting these
demonstrations to the held-out test input.

The resulting ARC evaluation is therefore transductive, following the protocol of
the prior recursive reasoning models we compare against.

\subsection{Evaluation Splits}

Table~\ref{tab:dataset_summary} summarizes the role of each domain in the evaluation.
Sudoku and Maze use their standard held-out splits.
ARC-AGI-1 and ARC-AGI-2 use the transductive protocol described above.
Game of Life and Arithmetic additionally provide explicit out-of-distribution splits
constructed to isolate different forms of generalization.

\begin{table}[H]
\centering
\caption{
Summary of the six evaluation domains.
Game of Life and Arithmetic provide explicit controlled distribution shifts;
the remaining domains are retained primarily for comparability with prior recursive
reasoning models.
}
\label{tab:dataset_summary}
\small
\begin{tabular}{lccc}
\toprule
Domain & Input structure & Primary capability & Controlled OOD \\
\midrule
Sudoku &
$9\times9$ grid &
Constraint satisfaction &
-- \\

Maze &
$30\times30$ grid &
Search and planning &
-- \\

Game of Life &
Variable-size binary grid &
Iterative dynamics &
Patterns, horizon \\

Arithmetic &
RPN expression &
Symbolic composition / search &
Operands, value range \\

ARC-AGI-1 &
Grid transformations &
Abstract rule induction &
-- \\

ARC-AGI-2 &
Grid transformations &
Abstract rule induction &
-- \\
\bottomrule
\end{tabular}
\end{table}
\FloatBarrier

\section{Training and Implementation Details}
\label{app:training-details}

This appendix provides implementation details omitted from
Section~\ref{sec:method}. The main text specifies the mechanisms needed to
define the model; here we report the corresponding optimizer,
regularization, and recurrent-carry settings.

\subsection{Backbone and Optimization}

The recurrent module consists of four post-norm Transformer layers with hidden
size 512 and eight attention heads. The feed-forward block uses SwiGLU with
expansion factor 4. We use rotary position embeddings and RMSNorm, and execute
the model in bfloat16.

The high- and low-level recurrent modules share parameters, giving 13.6M
trainable parameters in the configuration used for the main experiments.

We optimize the model with Adam-atan2. Training uses warm-up followed by either
cosine or exponential decay, depending on the domain configuration. Peak
learning rate, $\beta_2$, weight decay, and training duration are specified
separately for each domain in
Table~\ref{tab:domain-optimization-config}.

Puzzle embeddings are optimized separately using distributed sign-SGD. Their
learning rate is domain-specific and is reported in
Table~\ref{tab:domain-optimization-config}.

Gradients are clipped to a global norm of 1 after distributed reduction and,
when applicable, after accumulation. We record the pre-clipping norm as a
training-stability diagnostic.

We maintain an exponential moving average of all trainable parameters with
decay $0.999$, including the puzzle-embedding table.

\subsection{Recurrent-State Regularization}

\paragraph{Bounded relative update.}
For each token position, the relative size of the proposed state update is

\begin{equation}
r =
\frac{
\|\tilde z_L-z_L\|_2
}{
\|z_L\|_2+\epsilon
},
\qquad
\epsilon=10^{-8}.
\end{equation}

When enabled, we use $\tau=0.7$ and rescale the update as

\begin{equation}
\tilde z_L
\leftarrow
z_L +
\frac{
\tilde z_L-z_L
}{
\max(r/\tau,1)
}.
\end{equation}

The scaling coefficient is detached from the computation graph.

\paragraph{Update gate.}
The gate is initialized by scaling its weight vector by $0.1$ and setting its
bias to zero, giving $\alpha \approx 1/2$ at initialization.

\paragraph{Recurrence-consistent dropout.}
Dropout masks are sampled when an example enters an ACT trajectory and are
stored in the recurrent carry. The same masks are reused at every recurrent
cycle and every ACT step until that sequence halts.

Masks are sampled separately for the QKV projection, attention residual,
feed-forward residual, and feed-forward intermediate activations. Additional
masks are applied to the input embeddings and the two recurrent states; the
state masks are shared across token positions.

\paragraph{State and embedding noise.}
At the beginning of each ACT step during training, we perturb the input
embeddings and both recurrent states according to

\begin{equation}
u
\leftarrow
u + \eta \|u\|_2 \epsilon,
\qquad
\epsilon\sim\mathcal{N}(0,I),
\qquad
u\in\{x,z_H,z_L\}.
\end{equation}

The norm is computed per position and detached from the computation graph. We
use $\eta\in[0.003,0.005]$ depending on the domain.

\subsection{ACT and Recurrent Carry}

The model uses an adaptive-computation-time Q-head with a maximum budget of
16 ACT steps, reduced to 8 for ARC-AGI-1.

Recurrent states are initialized from learned vectors. At the end of an ACT
step, the resulting states are stored in the carry and detached before the
next ACT step. This allows later ACT steps to continue refining the same
trajectory without backpropagating through previous ACT steps.

The training objective is stablemax cross-entropy over the supervised output
positions together with the two halting losses.

At evaluation time, the learned halting decision is disabled for the main
accuracy comparison and every sequence executes the complete evaluation
budget, as described in Section~\ref{sec:main-results}.

\subsection{Domain-Specific Training Configurations}
\label{app:final-training-configurations}

We report the domain-specific configurations used for the six final models.
The recurrent architecture itself is fixed across domains, while
regularization and optimization hyperparameters are adjusted to the
characteristics of each task. Tables~\ref{tab:domain-architecture-config}
and~\ref{tab:domain-optimization-config} summarize these differences.

\paragraph{Shared configuration.}
All models use the shared recurrent Transformer described in
Section~\ref{sec:configuration}, with hidden size 512, four Transformer
layers, eight attention heads, and feed-forward expansion factor four.
Each ACT step performs
\((H_{\mathrm{cycles}},L_{\mathrm{cycles}})=(4,2)\) recurrent updates, and
training propagates gradients through the final
\((K_H,K_L)=(2,2)\) cycles of the trajectory.

The recurrent states \(z_H\) and \(z_L\) use post-cycle normalization,
recurrence-consistent dropout, and norm-scaled Gaussian perturbations.
Controlled \(z_L\) updates use the relative-update threshold
\(\tau=0.7\) together with the learned gate \(\alpha\).
The remaining domain-specific choices are the dropout rates, noise scale
\(\eta\), and whether convolutional input mixing and the \(z_L\) gate are
enabled.

\begin{table}[H]
  \centering
  \small
  \setlength{\tabcolsep}{5pt}
  \begin{tabular}{lrrrrr}
    \toprule
    Domain & \(d_{\mathrm{core}}\) & \(d_H=d_L\) & Conv. &
    \(\eta\) & \(z_L\)-gate \\
    \midrule
    Arithmetic   & .025 & .010 & no  & .005 & yes \\
    Sudoku       & .010 & .000 & no  & .005 & yes \\
    Game of Life & .100 & .000 & no  & .005 & yes \\
    Maze         & .010 & .000 & no  & .005 & yes \\
    ARC-AGI-1    & .025 & .025 & yes & .003 & yes \\
    ARC-AGI-2    & .025 & .025 & yes & .005 & yes \\
    \bottomrule
  \end{tabular}
  \caption{\textbf{Domain-specific recurrent regularization.}
  \(d_{\mathrm{core}}\) denotes dropout in the embedding, attention,
  residual, and feed-forward paths, while \(d_H\) and \(d_L\) denote
  dropout applied to the recurrent states \(z_H\) and \(z_L\).
  \(\eta\) is the relative state and embedding noise scale.
  ``Conv.'' indicates convolutional input mixing.
  Controlled \(z_L\) updates use \(\tau=0.7\).}
  \label{tab:domain-architecture-config}
\end{table}
\FloatBarrier

\paragraph{Optimization.}
The optimization procedure follows Section~\ref{sec:configuration}: all
models use Adam-atan2, global gradient clipping at norm 1, and EMA weights
for evaluation. The main domain-specific differences are the physical batch
size, training duration, learning-rate schedule, \(\beta_2\), weight decay,
and puzzle-embedding learning rate. Arithmetic uses a larger peak learning
rate and stronger weight decay, while Game of Life uses the largest physical
batch. Maze and ARC-AGI use substantially longer schedules with smaller
physical batches.

\FloatBarrier
\begin{table*}[!t]
  \centering
  \scriptsize
  \setlength{\tabcolsep}{5pt}
  \begin{tabular}{lrrrrrrrr}
    \toprule
    Domain & Physical batch & Epochs & Eval.\ interval & Peak LR &
    Min.\ LR ratio & \(\beta_2\) & Weight decay & Puzzle LR \\
    \midrule
    Arithmetic   & 4096 &   2000 &    50 & \(5{\times}10^{-4}\) &
      .01 & .95  & 1.0 & .005 \\
    Sudoku       & 4096 &   2000 &    50 & \(1{\times}10^{-4}\) &
      1.0 & .95  & 0.1 & .010 \\
    Game of Life & 8192 &   2000 &    50 & \(1{\times}10^{-4}\) &
      .10 & .95  & 0.1 & .010 \\
    Maze         & 1024 &  54000 &   500 & \(1{\times}10^{-4}\) &
      .01 & .995 & 0.1 & .010 \\
    ARC-AGI-1    &  768 & 300000 & 10000 & \(1{\times}10^{-4}\) &
      1.0 & .95  & 0.1 & .010 \\
    ARC-AGI-2    &  768 & 300000 & 10000 & \(1{\times}10^{-4}\) &
      1.0 & .95  & 0.1 & .010 \\
    \bottomrule
  \end{tabular}
  \caption{\textbf{Domain-specific optimization configurations.}
  All runs use 2,000 learning-rate warmup steps, \(\beta_1=0.9\),
  global gradient clipping at norm 1, and EMA weights.
  Puzzle embeddings use a separate learning rate (``Puzzle LR'')
  and the same weight decay as the remaining parameters.}
  \label{tab:domain-optimization-config}
\end{table*}
\FloatBarrier

\section{Evaluation and Baseline Details}
\label{app:evaluation-details}
\subsection{ARC-AGI Evaluation}
\label{app:arc-evaluation}

ARC-AGI is scored with the standard pass@$k$ protocol. For each test input,
predictions from all augmentations are first mapped back to the original
coordinate and color systems using the inverse spatial transformation and
inverse color permutation. Outputs are then cropped at the end-of-grid markers
and grouped by exact grid equality.

Candidate grids are ranked by vote count across augmentations, with the
halting-head confidence used as a tie-break. A puzzle is counted as solved at
pass@$k$ when the ground-truth output appears among the top-$k$ candidates for
each of its test inputs. We report pass@1 and pass@2.

The main evaluation disables adaptive halting and executes the full ACT budget
for every sequence, keeping batches synchronized and removing halting errors
from the quality comparison. Test-time-scaling experiments instead extend the
ACT budget and compare the resulting predictions.
\subsection{Inference Compute}
\label{app:compute}

A small parameter count does not imply proportionally small inference cost in
a recursive model. Weight sharing reduces parameter storage, but the shared
block is still executed repeatedly.

For a model with $H_{\mathrm{cycles}}$ high-level cycles,
$L_{\mathrm{cycles}}$ low-level cycles, $H_{\mathrm{layers}}$ high-level
layers, and $L_{\mathrm{layers}}$ low-level layers, one ACT step requires

\begin{equation}
H_{\mathrm{cycles}}
\left(
L_{\mathrm{cycles}}L_{\mathrm{layers}}
+
H_{\mathrm{layers}}
\right)
\end{equation}

layer applications. If an example executes $T$ ACT steps, this cost is
multiplied by $T$.

\begin{table}[H]
\centering
\caption{
Inference cost of one answer relative to a single forward pass of the
non-recurrent dense control. $T$ denotes the number of ACT steps.
}
\label{tab:compute}
\begin{tabular}{lccc}
\toprule
Model & Layers $(H,L)$ & Cycles $(H,L)$ & Relative compute \\
\midrule
Dense Transformer & $8,-$ & -- & $1$ \\
HRM & $4,4$ & $(2,2)$ & $3T$ \\
TRM & $2,2$ & $(3,6)$ & $5.25T$ \\
URM & $2,2$ & $(3,6)$ & $5.25T$ \\
Ours & $4,4$ & $(4,2)$ & $6T$ \\
\bottomrule
\end{tabular}
\end{table}
\FloatBarrier

Parameter sharing therefore primarily reduces parameter count rather than
sequential computation: TRM and our configuration have substantially different
parameter counts but similar recurrent inference cost.

The dense Transformer is much cheaper per answer and is therefore used as a
control for recurrence rather than as a compute-matched baseline. On ARC-AGI,
the augmentation-and-voting protocol introduces an additional inference cost
because each puzzle is evaluated over many transformed versions. General-purpose
LLMs are omitted from Table~\ref{tab:compute}, since their inference cost is
determined by a different autoregressive computation pattern.

\section{Gradient-Path Length Ablation}
\label{app:gradient-path-length}

Our hierarchical recurrence contains four high-level cycles and two low-level
cycles per high-level update.  The default implementation truncates
backpropagation through this recurrence.  We ablate the number of terminal
high-level and low-level cycles retained in the backward graph, denoted by
\(g_H\in\{1,2,3,4\}\) and \(g_L\in\{1,2\}\), respectively.  This produces a
complete \(2\times4\) set of evaluated settings.  The forward recurrence,
inference cost, and parameter count are unchanged; the principal architectural
difference is the path through which gradients are propagated.

\subsection{Setup.}
All runs use the same 13.65M-parameter Arithmetic model: hidden size 512,
four layers in each hierarchy, eight attention heads, four high-level cycles,
two low-level cycles, and at most 16 ACT steps.  They run for 2000 epochs
(439,600 optimizer updates) with global batch size 4096, EMA weights, and
seed 125.

Evaluation is performed every 50 epochs on the in-distribution range
\(0\)--\(101\) and the extrapolation range \(102\)--\(201\).  We report the
EMA \texttt{best\_full} exact score over the full recorded evaluation
trajectory. 

\begin{figure}[t]
  \centering
  \includegraphics[width=\linewidth]{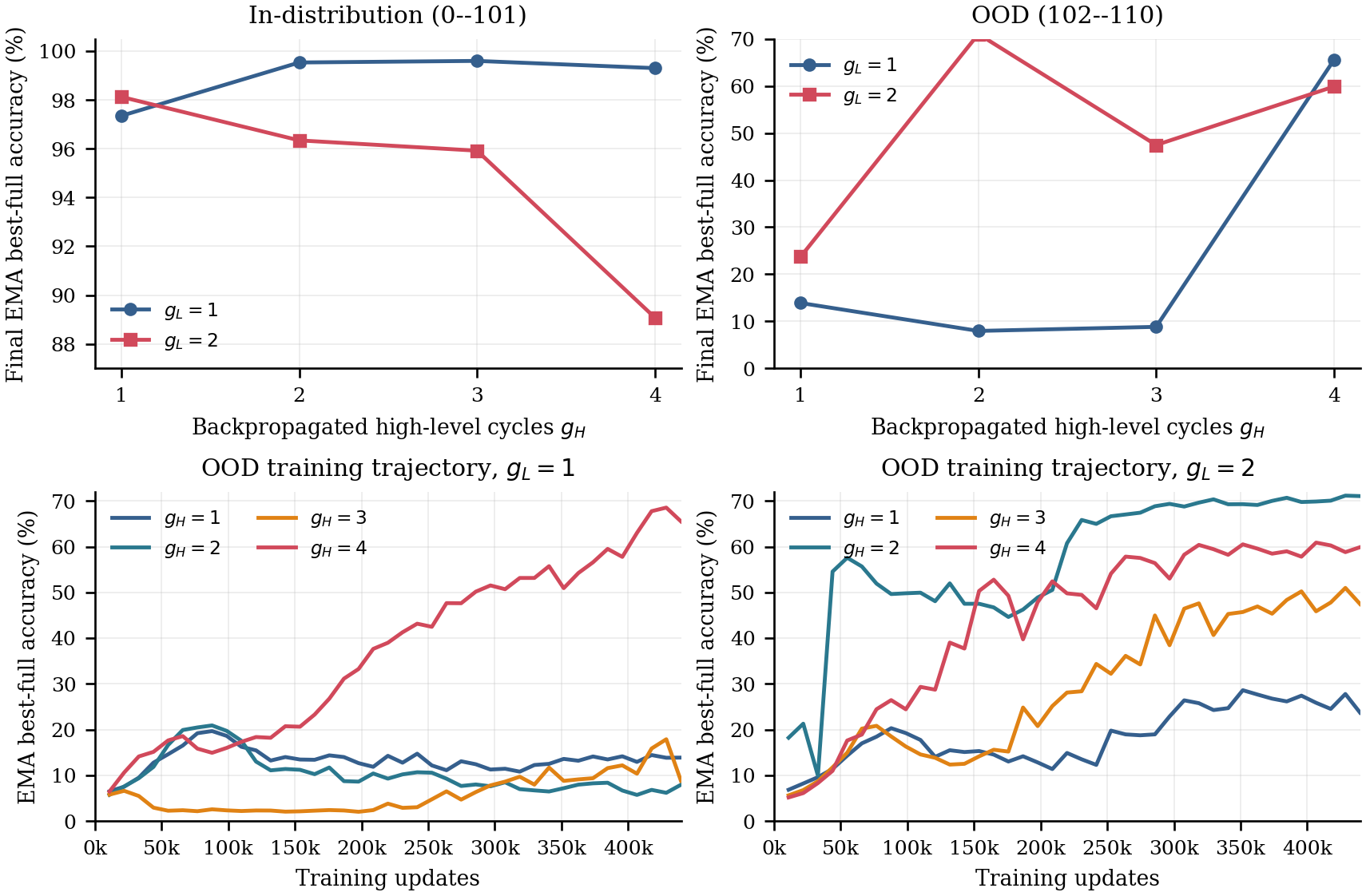}
  \caption{\textbf{Effect of the gradient-path length on Arithmetic.}
  Top: final EMA \texttt{best\_full} exact accuracy as a function of the
  number of high-level cycles retained in the backward graph, for one or two
  retained low-level cycles.  Bottom: OOD accuracy over training.  The
  selected \((2,2)\) model obtains the highest final OOD score.}
  \label{fig:gradient-path-length}
\end{figure}

\begin{table}[H]
  \centering
  \small
  \setlength{\tabcolsep}{5pt}
  \begin{tabular}{rrrrrr}
    \toprule
    \(g_L\) & \(g_H\) & Final ID & Final OOD & Best OOD &
    Best update \\
    \midrule
    1 & 1 & 97.35 & 13.87 & 19.67 &  87,920 \\
    1 & 2 & 99.54 &  7.93 & 20.91 &  87,920 \\
    1 & 3 & \textbf{99.60} &  8.80 & 17.88 & 428,610 \\
    1 & 4 & 99.31 & 65.53 & 68.54 & 428,610 \\
    2 & 1 & 98.12 & 23.71 & 28.61 & 351,680 \\
    2 & 2 & 96.34 & \textbf{71.16} &
    \textbf{71.16} & 428,610 \\
    2 & 3 & 95.92 & 47.46 & 50.98 & 428,610 \\
    2 & 4 & 89.08 & 59.88 & 60.88 & 406,630 \\
    \bottomrule
  \end{tabular}
  \caption{\textbf{EMA \texttt{best\_full} exact accuracy under gradient
  truncation.}  Accuracies are percentages.  Final values are measured at
  439,600 updates.  ``Best OOD'' is the maximum over the 40 recorded
  checkpoints and uses the OOD split for checkpoint selection.}
  \label{tab:gradient-path-length}
\end{table}
\FloatBarrier

\subsection{The selected \((2,2)\) setting balances fitting and generalization.}
The \((g_L,g_H)=(2,2)\) run reaches \(71.16\%\) final OOD accuracy and
\(71.16\%\) at its best checkpoint, the highest values in the comparison,
while retaining \(96.34\%\) ID accuracy.  Its final OOD score exceeds the
next-best \((1,4)\) setting by 2.62 percentage points.  In contrast, several
shorter paths reach \(97.35\%\)--\(99.60\%\) ID accuracy but remain below
\(14\%\) OOD, a pattern consistent with fitting in-range regularities without
learning an extrapolating procedure.  At the other extreme, extending both
paths to \((2,4)\) reduces ID to \(89.08\%\) and OOD to \(59.88\%\), indicating
that additional backpropagation depth does not translate monotonically into
better optimization.

\subsection{An intermediate gradient path is optimal.}
The results expose a three-way trade-off.  Paths that are too short provide
insufficient long-range credit assignment and favor an ID--OOD generalization
gap; paths that are too long can make optimization harder without improving
the learned computation.  The intermediate \((2,2)\) configuration provides
enough gradient reach to learn the transferable arithmetic procedure while
avoiding the degradation observed at the longest setting.  It is therefore
the best observed balance between in-distribution fitting and extrapolative
generalization, rather than simply the model with the greatest backward
depth.

\section{Game of Life: Single-Seed Training Dynamics}
\label{app:game-of-life-training}

This appendix examines one seed of the final Game of Life configuration to
illustrate the dynamics of training.  It is not used as a seed-averaged result
or as an additional model comparison.  The evaluation contains an
in-distribution (ID) split, four out-of-distribution (OOD) horizon shifts
(\(+1,+2,+3,+10\) Game of Life steps), and an OOD split with unseen initial
patterns.

\paragraph{Setup.}
The model has 13.65M parameters, hidden size 512, four layers in each
hierarchy, eight attention heads, four high-level cycles, two low-level
cycles, and at most 16 ACT steps in the training configuration.  Gradients
are retained through the last two high- and two low-level cycles.  The model
uses RoPE, a gated low-level state, stablemax cross-entropy, and dropout 0.1
on embeddings, attention, residual, and feed-forward paths.

Training uses a global batch size of 8192 across a world size of 16, peak
learning rate \(10^{-4}\) after 2000 warmup updates, minimum learning-rate
ratio 0.1, weight decay 0.1, EMA parameters, and seed 536.  The nominal
schedule is 2000 epochs, with evaluation and checkpointing every 50 epochs.
The scalar history reaches epoch 1999 and includes the complete epoch-2000
evaluation at 367,200 optimizer updates.  The job was terminated after this
evaluation rather than being registered as completed, so the full planned
training trajectory is nevertheless available.

\begin{figure}[t]
  \centering
  \includegraphics[width=\linewidth]{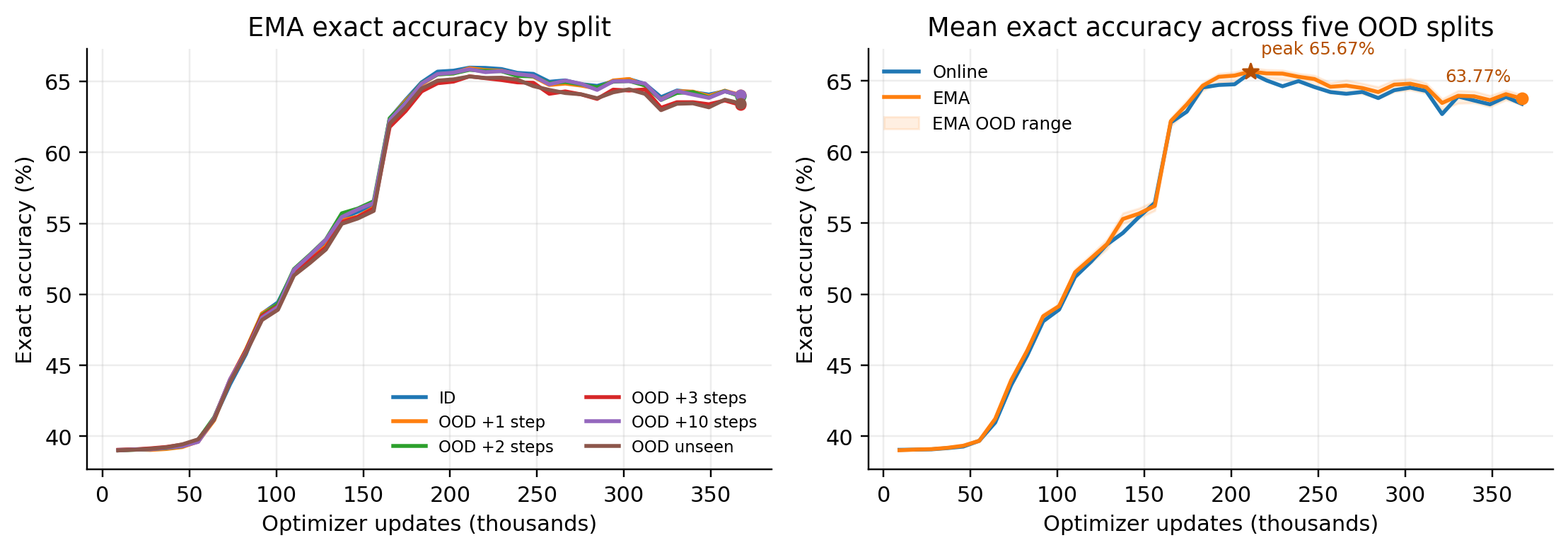}
  \caption{\textbf{Single-seed Game of Life training trajectory.}
  Left: EMA \texttt{best\_full} exact accuracy on ID and five OOD splits.
  Right: the unweighted OOD mean for online and EMA parameters; shading spans
  the minimum and maximum EMA OOD split.  Accuracy peaks at epoch 1150 and
  declines before the final checkpoint.}
  \label{fig:game-of-life-training}
\end{figure}

\FloatBarrier
\begin{table*}[!t]
  \centering
  \scriptsize
  \setlength{\tabcolsep}{5pt}
  \begin{tabular}{lrrrrrr}
    \toprule
    Split & Online final & Online peak & EMA final & EMA peak &
    EMA token acc. & Best forced step \\
    \midrule
    ID                  & 63.70 & 65.99 & 64.01 & 65.98 & 75.81 & 63.75 (4) \\
    OOD \(+1\) step     & 63.71 & 65.73 & 64.04 & 65.94 & 75.83 & 63.79 (4) \\
    OOD \(+2\) steps    & 63.65 & 65.73 & 63.94 & 65.83 & 75.92 & 63.67 (3) \\
    OOD \(+3\) steps    & 63.02 & 65.22 & 63.35 & 65.37 & 75.45 & 63.03 (3) \\
    OOD \(+10\) steps   & 63.46 & 65.94 & 64.04 & 65.87 & 75.92 & 63.76 (4) \\
    OOD unseen patterns & 63.12 & 65.34 & 63.46 & 65.36 & 75.69 & 63.12 (5) \\
    \bottomrule
  \end{tabular}
  \caption{\textbf{Final and peak validation performance.}
  Values are percentages.  Exact columns use the logged
  \texttt{best\_full} exact-accuracy series; token accuracy is the EMA
  \texttt{accuracy} series at the final checkpoint.  All EMA peaks occur at
  211,140 updates (epoch 1150).  ``Best forced step'' reports the highest single-step exact
  score at the final checkpoint, followed by its logged ACT step in
  parentheses.}
  \label{tab:game-of-life-final}
\end{table*}
\FloatBarrier

\paragraph{The best checkpoint occurs well before the end of training.}
The ID EMA exact score rises from 39.04\% at epoch 50 to 48.64\% at epoch 500
and 64.94\% at epoch 1000.  It peaks at 65.98\% at epoch 1150, then declines
to 64.01\% at epoch 2000.  The five-split OOD mean follows the same trajectory:
39.02\%, 48.44\%, 64.67\%, a peak of 65.67\%, and a final value of 63.77\%.
Selecting the final checkpoint would therefore understate the best observed
ID and OOD results by 1.97 and 1.91 percentage points, respectively.

\paragraph{The measured ID--OOD gap is small.}
At the best checkpoint, ID exact accuracy is 65.98\% and the OOD mean is
65.67\%, a difference of 0.31 percentage points; the OOD range is
65.36\%--65.94\%.  At the final checkpoint, the corresponding values are
64.01\% and 63.77\%, with an OOD range of 63.35\%--64.04\%.  The \(+3\)
horizon shift is the hardest final split, while \(+1\) and \(+10\) are
marginally above ID.  The differences remain small and non-monotonic.

\begin{figure}[t]
  \centering
  \includegraphics[width=0.86\linewidth]{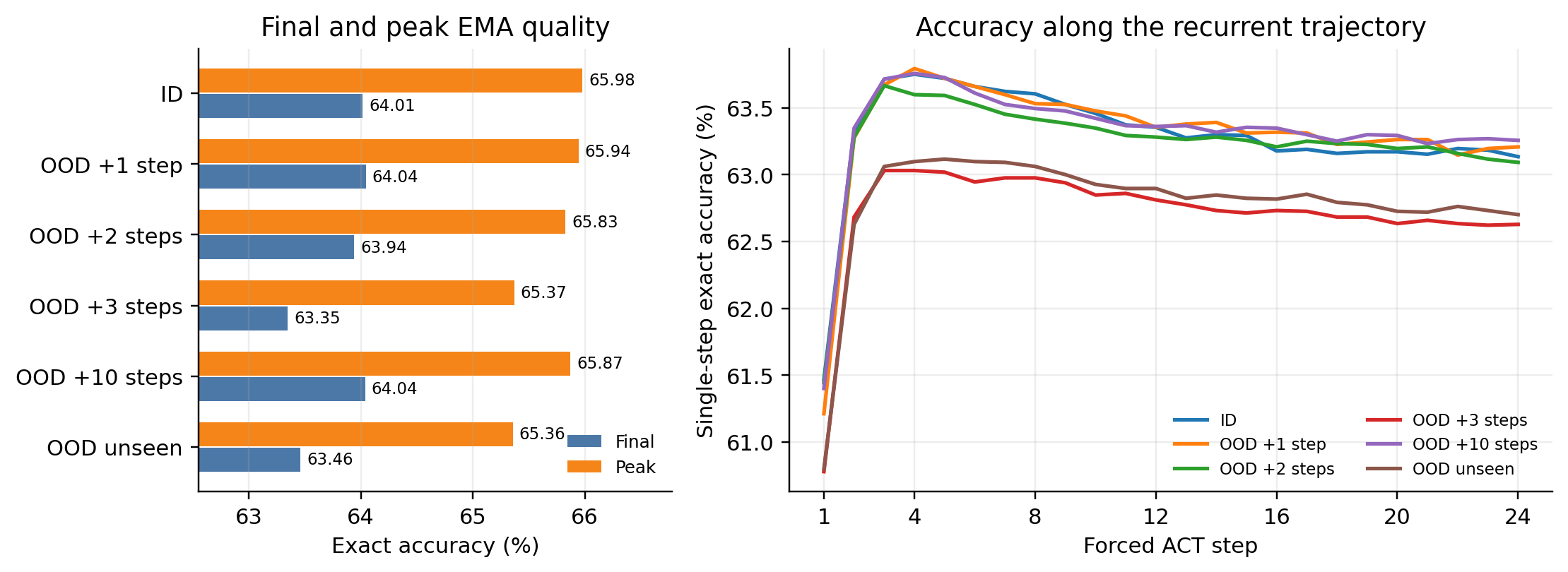}
  \caption{\textbf{Final and peak EMA \texttt{best\_full} quality and
  recurrent-depth profile.}  Left: final and peak \texttt{best\_full} exact
  accuracy for every evaluation split.  Right: EMA single-step exact accuracy at the final
  complete checkpoint.  Most of the single-step gain occurs between forced
  ACT steps 1 and 2; all six curves then remain in a narrow band through
  step 24.}
  \label{fig:game-of-life-final-depth}
\end{figure}

\paragraph{Additional recurrence is not used for progressive refinement.}
At the final checkpoint, exact accuracy jumps by roughly two points over the
first few forced steps.  The best single-step result occurs at step 3, 4, or
5 depending on the split, after which performance is flat or drifts slightly
downward.  This depth profile indicates that the final model produces nearly
all of its useful refinement at the start of the recurrent trajectory; the
logged continuation to step 24 does not yield a second phase of computation.

\paragraph{Metric scope and limitations.}
This run is one of the seeds of the final Game of Life configuration and is
shown only to characterize training dynamics.  Conclusions about final model
quality should rely on the seed-aggregated results in the main paper.  This
single trajectory does not support a causal comparison between architectures
or optimization methods.

\section{A Negative Result: Joint Training across Four Domains}
\label{app:multitask-training}

We also tested whether a single model could learn Arithmetic, Sudoku, Game of
Life, and Maze jointly.  To remove domain-specific differences at the input
and output interfaces, we converted all four datasets to the same binary
serialization and used a shared tokenizer.  The resulting examples were
combined into one training corpus, and no domain-specific parameters or
prediction heads were introduced.

\paragraph{Setup.}
The joint model has 27.3M parameters, hidden size 512, four layers in each
hierarchy, eight attention heads, and \(H=L=8\) recurrent cycles.  We used a
global batch size of 8192, a peak learning rate of \(10^{-5}\), weight decay
of 1.0, a maximum of 16 ACT steps, and an exponential moving average (EMA) of
the parameters.  The run used seed 125.  Although the configured schedule was
2000 epochs, the ClearML task was stopped after epoch 300 (195,093 optimizer
updates).  We therefore report all six available validation checkpoints,
recorded every 50 epochs.

\begin{figure}[t]
  \centering
  \includegraphics[width=\linewidth]{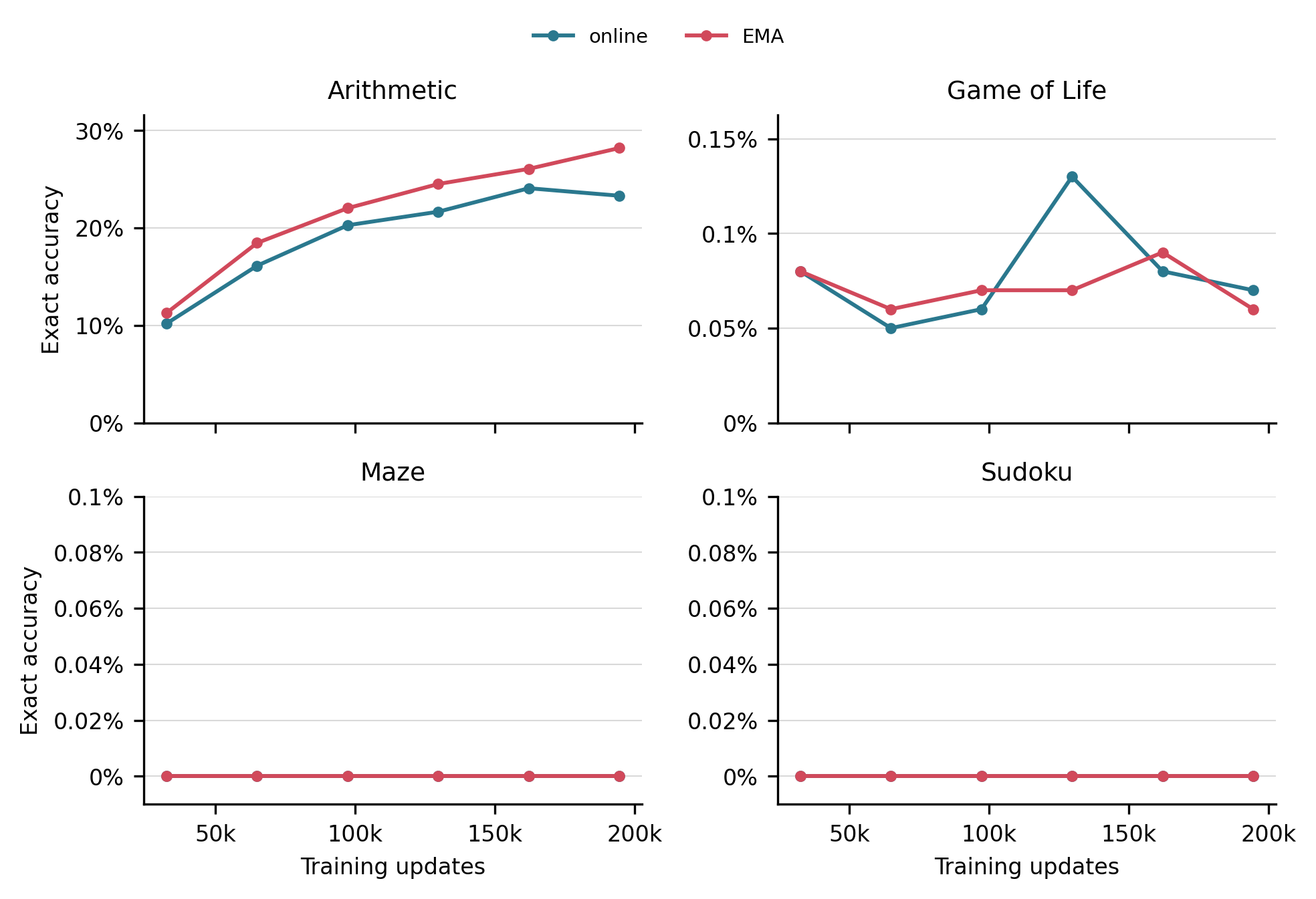}
  \caption{\textbf{Exact validation accuracy during joint four-domain
  training.}  Online and EMA parameters are evaluated at six checkpoints.
  Arithmetic improves throughout training, but Game of Life remains below
  0.1\% exact accuracy and neither Maze nor Sudoku produces a single exact
  solution.  The panels use different vertical scales.}
  \label{fig:multitask-exact-accuracy}
\end{figure}

\begin{table}[H]
  \centering
  \small
  \setlength{\tabcolsep}{6pt}
  \begin{tabular}{lrrr}
    \toprule
    Domain & Final token acc. & Best exact acc. & Final exact acc. \\
    \midrule
    Arithmetic   & 63.97 & 28.22 & 28.22 \\
    Game of Life & 51.68 &  0.09 &  0.06 \\
    Maze         & 87.50 &  0.00 &  0.00 \\
    Sudoku       & 20.21 &  0.00 &  0.00 \\
    \bottomrule
  \end{tabular}
  \caption{\textbf{EMA validation metrics for joint training.}  Values are
  percentages.  ``Best'' is selected over the six checkpoints; final metrics
  are measured at 194,562 updates.  Token accuracy can be high even when the
  full structured output is always incorrect.}
  \label{tab:multitask-training}
\end{table}
\FloatBarrier

\paragraph{Outcome.}
The shared model learns only a partial Arithmetic solver.  Arithmetic EMA
exact accuracy increases from 11.27\% at the first checkpoint to 28.22\% at
the last.  In contrast, Game-of-Life exact accuracy fluctuates between
0.06\% and 0.09\% for the EMA model, while Maze and Sudoku remain exactly
zero at every checkpoint.  The discrepancy between token and exact accuracy
is especially pronounced for Maze: the final token accuracy is 87.50\%, yet
none of the complete predicted paths is correct.  Thus the token-level loss
can improve by matching many local symbols without learning the global
constraints required for a valid solution.

\paragraph{Interpretation and limitations.}
Unifying the representation is therefore not sufficient, within the observed
compute budget, to obtain a useful universal solver.  The result is
consistent with severe cross-domain optimization interference or with the
joint objective being dominated by easy token-level regularities.  This run
does not distinguish these explanations from data-mixture imbalance,
insufficient model capacity, or an inadequate optimization schedule.
Moreover, the task was stopped before its nominal 2000-epoch schedule and was
run with a single seed; no error is present in the recorded log that would
identify why it was stopped.  We consequently treat this experiment as a
negative diagnostic result rather than evidence that multitask training is
intrinsically ineffective, and we use separately trained domain models in
the main experiments.

\end{document}